\RequirePackage{fix-cm}
\documentclass[smallcondensed]{svjour3}     \smartqed  \journalname{Machine Learning}
\makeatletter
\let\cl@chapter\undefined
\makeatletter
\usepackage{natbib}
\usepackage{graphicx}
\usepackage{amsmath,amssymb}
\usepackage{color}
\usepackage{colortbl}
\usepackage{tikz}
\usepackage[ruled,linesnumbered,vlined]{algorithm2e}
\usepackage[misc,geometry]{ifsym}
\usepackage[caption=false]{subfig}
\usepackage[hidelinks]{hyperref} 
\usepackage[capitalize]{cleveref}
\Crefname{line}{line}{lines}
\newcommand{\lzero}{$\ell_0$}
\newcommand{\lzeronorm}{\lzero{}-``norm''}
\newcommand{\lone}{$\ell_1$}
\newcommand{\lonenorm}{\lone{}-norm}
\DeclareMathOperator{\vecop}{vec}
\DeclareMathOperator{\argmin}{argmin}
\DeclareMathOperator{\argmax}{argmax}
\definecolor{brightpink}{rgb}{1.0, 0.0, 0.5}

\newcommand{\revise}[1]{#1}

\newcommand{\eproof}{\space
    {\ \vbox{\hrule\hbox{\vrule height1.3ex\hskip0.8ex\vrule}\hrule}}\par}

\begin{document}

\title{Matrix-wise $\ell_0$-constrained Sparse Nonnegative\\ Least Squares
\thanks{NN and NG acknowledge the support by the European Research Council (ERC starting grant No 679515), and by the Fonds de la Recherche Scientifique - FNRS and the Fonds Wetenschappelijk Onderzoek - Vlanderen (FWO) under EOS project O005318F-RG47. NG also acknowledges the Francqui foundation. JEC acknowledges the support of the ANR grant ANR JCJC LoRAiA ANR-20-CE23-0010.}
}

\titlerunning{Matrix-wise $\ell_0$-constrained Sparse NNLS}

\author{Nicolas Nadisic$^{1}$         \and
        Jeremy E. Cohen$^{2}$         \and
        Arnaud Vandaele$^{1}$         \and
        Nicolas Gillis$^{1}$
}

\authorrunning{N. Nadisic et al.} 

\institute{\Letter{}~Nicolas Nadisic \\
           nicolas.nadisic@umons.ac.be\\
           $^{1}$ Department of Mathematics and Operational Research, University of Mons, Mons, Belgium
           $^{2}$ Univ Lyon, INSA-Lyon, UCBL, UJM-Saint Etienne, CNRS, Inserm,
CREATIS UMR 5220, U1206, F-69100 Villeurbanne, France
}

\date{Received: date / Accepted: date}

\maketitle

\begin{abstract}

\color{white} i \color{black} 

Nonnegative least squares problems with multiple right-hand sides (MNNLS) arise in models that rely on additive linear combinations. In particular, they are at the core of most nonnegative matrix factorization algorithms and have many applications. 
The nonnegativity constraint is known to naturally favor sparsity, that is, solutions with few non-zero entries.
However, it is often useful to further enhance this sparsity, as it improves the interpretability of the results and helps reducing noise, which leads to the sparse MNNLS problem.   
In this paper, as opposed to most previous works that enforce sparsity column- or row-wise, we first introduce a novel formulation for sparse MNNLS, with a matrix-wise sparsity constraint.
Then, we present a two-step algorithm to tackle this problem. 
The first step divides sparse MNNLS in subproblems, one per column of the original problem. 
It then uses different algorithms to produce, either exactly or approximately, a Pareto front for each subproblem, that is, to produce a set of solutions representing different tradeoffs between reconstruction error and sparsity. 
The second step selects solutions among these Pareto fronts in order to build a sparsity-constrained matrix that minimizes the reconstruction error. 
We perform experiments on facial and hyperspectral images, and we show that our proposed two-step approach provides more accurate results than state-of-the-art sparse coding heuristics applied both column-wise and globally.

\end{abstract}

\section{Introduction}
\label{sec:intro}

Nonnegative least squares (NNLS) problems arise in many applications where data points can be represented as additive linear combinations of meaningful components \citep{lee1997unsupervised}.
For instance, 
\begin{itemize}

\item In facial images, the faces are the nonnegative linear combination of facial features such as eyes, noses and lips~\citep{lee1999learning}. 

    \item In hyperspectral images, the spectral signature of a pixel is the nonnegative linear combination of the spectral signature of the materials it contains \citep{bioucas2012hyperspectral}. 
    
\end{itemize}
NNLS problems are also at the core of most approaches to solve nonnegative matrix factorization (NMF); see \cite[Chapter 8]{gillis2020} and the references therein.  
The standard NNLS problem can be formulated as follows: 
given a dictionary matrix $A \in \mathbb{R}^{m \times r}$ and a data vector $b \in \mathbb{R}^m$, solve
\begin{equation}\label{eq:nnls}
    \min\limits_{x} \| Ax - b \|_2^2 
    \quad \text{ such that } \quad x \geq 0 . 
\end{equation}
\revise{Note that \eqref{eq:nnls} is a convex problem.}

\subsection{Sparsity and NNLS}\label{sec:intronnls}

The nonnegativity constraint is known to naturally produce sparse solutions, that is, solutions with few non-zero entries~\citep{foucart2014sparse}.
Sparsity often improves the interpretability of the results by modelling data points as combinations of only a few components.
For example, in hyperspectral unmixing, that is, the task of identifying materials in a hyperspectral image, sparsity means that a pixel contains only a few materials.

A natural sparsity measure is the \lzeronorm{}, defined as the number of non-zero entries in a given vector, $\| x \|_0 = | \{i : x_i \neq 0 \} |$. Given a positive integer $k$, a vector $x$ is said \revise{to be} $k$-sparse if $\| x \|_0 \leq k$. 

Unfortunately, the sparsity of the solution to \revise{an} NNLS problem is not guaranteed in general, whereas controlling it can be helpful in many applications. 
For this reason, numerous techniques have been developed to favor sparsity.

A sparsity-constrained variant of Problem~(\ref{eq:nnls}), referred to as \emph{$k$-sparse NNLS}, is the following 
\begin{equation}\label{eq:ksnnls}
    \min\limits_{x} \| Ax - b \|_2^2 \quad \text{ such that } \quad  x \geq 0 \; \text{ and } \; \| x \|_0 \leq k.
\end{equation}
Several algorithms exist to tackle Problem \eqref{eq:ksnnls}, either exactly or approximately; we detail them in \cref{sec:relwork}.

In hyperspectral unmixing, this $k$-sparsity constraint implies that a pixel can be composed of at most $k$ materials.
Although this formulation is intuitive, in some cases setting the parameter $k$ is not straightforward.
Therefore, we can also consider a \emph{biobjective} formulation where the objectives are, on the one hand, to minimize the reconstruction error, and on the other hand, to maximize the sparsity (that is, minimize the \lzeronorm{}),
\begin{equation}\label{eq:biobj}
    \min\limits_{x \geq 0} \{ \| Ax - b \|_2^2  , \| x \|_0 \}.
\end{equation}
As sparser solutions lead to higher error, these objectives are conflicting, so there is not an optimal solution to Problem \eqref{eq:biobj} and we need a trade-off between the two objectives.
Thus, we seek \emph{Pareto-optimal} solutions.

\begin{figure}[htb]
  \centering
  \definecolor{mypurple}{rgb}{0.5961,0.3059,0.6392}\definecolor{mygreen}{rgb}{0.7020,0.8706,0.4118}\definecolor{myorange}{rgb}{0.9843,0.5020,0.4471}

\begin{tikzpicture}[scale=1]

\coordinate (A) at (0,2.7);
\coordinate (B) at (1,2);
\coordinate (C) at (2,1.7);
\coordinate (D) at (3,1);
\coordinate (E) at (4,0.5);
\coordinate (F) at (5,0.5);

\draw[->] (0,0) node[anchor=north] {$0$} -- (6,0) node[anchor=west] {\small $\| x \|_0$};
\draw[->] (0,0) node[anchor=east] {$0$} -- (0,3) node[anchor=south, rotate=90, xshift=-16mm] {\small $\| Ax-b \|_2^2$};

\draw[dotted] (1,0) -- (B)
              (2,0) -- (C)
              (3,0) -- (D)
              (4,0) -- (E)
              (5,0) -- (F);
              
\draw  (1,0) node[below] {$1$};
\draw  (2,0) node[below] {$2$};
\draw  (3,0) node[below] {$3$};
\draw  (4,0) node[below] {$4$};
\draw  (5,0) node[below] {$r=5$};

\filldraw[thick] (A) circle (0.05) node[anchor=west] {\small $x = 0$};
\draw (A) node[left]{\small $\|b\|_2^2$};
\filldraw[thick] (B) circle (0.05);
\filldraw[thick] (C) circle (0.05);
\filldraw[thick] (D) circle (0.05);
\filldraw[thick] (E) circle (0.05);
\filldraw[thick] (F) circle (0.05) node[anchor=south] {\small $x \in\argmin\limits_{x \geq 0} \|Ax-b\|_2^2$};

\end{tikzpicture}   \caption{Example of the Pareto front for a biobjective $k$-sparse NNLS problem with $r = 5$ variables. 
  The first solution, for $\|x\|_0 = 0$, corresponds to the zero vector.
  The last solution, for $\|x\|_0 = 5$, corresponds to the NNLS problem with no sparsity constraint.
  Here the penultimate solution is identical to the last one, meaning that the solution with no sparsity constraint has naturally 1 zero entry.}
  \label{fig:paretofront}
\end{figure}
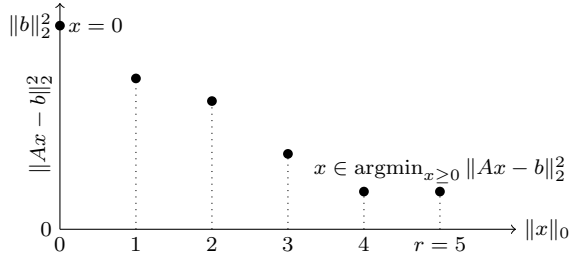

Given different objectives to optimize, a solution $x$ is said \revise{to be} Pareto-optimal if there does not exist any solution which is at least as good as $x$ on all objectives and strictly better than $x$ on at least one objective.
The set of all Pareto-optimal solutions for a given problem is called the \emph{Pareto front}, see \cref{fig:paretofront}.

Here, the discreteness of the \lzeronorm{} implies that solving Problem~\eqref{eq:biobj} conceptually reduces to solving 
Problem~\eqref{eq:ksnnls} for all possible values of $k$.
To the best of our knowledge, there exist only one algorithm to solve Problem~\eqref{eq:biobj} exactly \citep{nadisic2021biobjective}.
We will present it in \cref{subsec:arbo}.
It is also possible to modify some algorithms originally intended for $k$-sparse NNLS so that they generate an approximation of the Pareto front; we will see examples with greedy algorithms and the homotopy algorithm, respectively in \cref{subsec:greedy,subsec:homotopy}.

\subsection{Sparsity in NNLS problems with multiple right-hand sides} \label{sec:sparMMNLS}

In many cases, one has to deal with NNLS problems with multiple right-hand sides (MNNLS), that is, problems of the form
\begin{equation}\label{eq:multiplennls}
    \min\limits_{X} \| B - AX \|_F^2 \quad \text{ such that } \quad X \geq 0,
\end{equation}
where  $B \in \mathbb{R}^{m \times n}$, $A \in \mathbb{R}^{m \times r}$, and $X \in \mathbb{R}^{r \times n}$.
\revise{Given a matrix $B \in \mathbb{R}^{m \times n}$, we note $\|B\|_F$ its Frobenius norm, that is $\|B\|_F = \sqrt{\sum_{i=1}^m \sum_{j=1}^n B(i,j)^2 }$ where $B(i,j)$ is the entry of $B$ at position $(i,j)$.}
We note $B(:,j)$ the $j$th column of the matrix $B$.
Problem (\ref{eq:multiplennls}) can be decomposed into $n$ NNLS subproblems of the form (\ref{eq:nnls}), where $B(:,j)$, $A$, and $X(:,j)$ correspond to $b$, $A$, and $x$, respectively.  
For example, in the unmixing of a hyperspectral image, every column $B(:,j)$ represents a pixel, and the corresponding column $X(:,j)$ represents its composition, in terms of the abundances of the $r$ materials whose spectral signatures are the columns of $A$.
Note that in this work, for the sake of conciseness, we focus only on the (sparse) optimization of $X$, but all concepts and algorithms can be applied symmetrically on $A$.

This is closely related to the nonnegative matrix factorization (NMF) problem, of the form \begin{equation}\label{eq:nmf}
    \min\limits_{A, X}  \| B - AX \|_F^2 
    \quad \text{ such that } \quad   
    A \geq 0 \text{ and } X \geq 0,
\end{equation}
in which we aim to find the factors $A$ and $X$, given $B$ and a factorization rank~$r$.
The usual optimization scheme for NMF consists in alternatively optimizing one factor while fixing the other, which is equivalent to solving MNNLS subproblems.
Note that in this paper, we focus on MNNLS rather than NMF.

To encourage sparsity in MNNLS, one can apply a sparse NNLS model column-wise, leading to 
\begin{equation}\label{eq:ksmnnls}
    \min\limits_{X} \| B - AX \|_F^2 
    \quad \text{ such that } \quad X \geq 0 \text{ and } \| X(:,j) \|_0 \leq k \text{ for all } j. 
\end{equation}
Solving Problem \eqref{eq:ksmnnls} boils down to solving $n$ independent subproblems of the form \eqref{eq:ksnnls}. 
However, in some applications, setting the sparsity parameter $k$ is tricky, as the relevant value can vary for different columns.
For example, in hyperspectral unmixing, pixels will be composed of different numbers of materials. 
Therefore, one can consider a more global approach, such as
\begin{equation}\label{eq:matwisemnnls}
    \min\limits_{X} \| B - AX \|_F^2 
    \quad \text{ such that } \quad X \geq 0 \text{ and } \| X \|_0 \leq q,
\end{equation}
where $\| X \|_0 = \sum_j \|  X(:,j) \|_0$ and 
$q$ is a matrix-wise sparsity parameter, hence enforcing an  \emph{average} sparsity $q/n$ for the columns of $X$.
In the following, Problem~\eqref{eq:matwisemnnls} is called the matrix-wise $q$-sparse MNNLS problem, and solving it is the main focus of this paper.

Note that \eqref{eq:matwisemnnls} could theoretically be solved by any column-wise $k$-sparse NNLS algorithm, because \eqref{eq:matwisemnnls} is equivalent to the vectorized form
\begin{equation}\label{eq:mnnlsvec}
 \min_{\bar{x}} \| \vecop(B) - \underbrace{(A \otimes I)}_{\Omega} \bar{x} \|_2^2 \text{ such that } \bar{x} \geq 0 \text{ and } \|\bar{x}\|_0 \leq q, 
\end{equation}
where $\otimes$ is the Kronecker product, $I$ is the identity matrix of appropriate dimension, and
$\vecop(B)$ denotes the column vector obtained by stacking the columns of $B$ on top of one another.
Problem \eqref{eq:mnnlsvec} is a $k$-sparse NNLS problem, but in practice its dimensions make it difficult to solve directly.
Denoting $\Omega = A \otimes I$, we have $\Omega \in \mathbb{R}^{(mn) \times (rn)}$, which is particularly problematic in hyperspectral unmixing where the dimension $n$ can reach tens of thousands.

It is possible to implement some $k$-sparse NNLS algorithms in a non-naive way to solve \eqref{eq:mnnlsvec} efficiently without actually allocating $\Omega$, and we detail such implementation of a greedy algorithm in \cref{subsec:nnompg}.
However, even in this case, when $n$ is large then the problem to solve is huge and the computing time can become too high, see \cref{sec:xp} for an experimental illustration.

\subsection{Contribution and outline of the paper} 

The main goal of this work is to describe a novel method able to solve efficiently the matrix-wise $q$-sparse MNNLS problem \eqref{eq:matwisemnnls}, even in large dimensions.
This method can be summarized by two main steps:
\begin{enumerate}
    \item  Problem \eqref{eq:matwisemnnls} is divided in $n$ subproblems of the form \eqref{eq:biobj} and, for each of them, the Pareto front is computed with existing algorithms.\item One solution per column (hence per Pareto front) is selected to build a solution to Problem \eqref{eq:matwisemnnls}, that is, a $q$-sparse matrix.
    This combinatorial step is solved exactly with a dedicated algorithm.
\end{enumerate}
To the best of our knowledge, this work is the first to tackle specifically Problem~\eqref{eq:matwisemnnls}.
Note that the algorithms used in the first step are not original contributions.
The contributions lie rather in the use of these existing algorithms to generate the Pareto fronts of the subproblems, and the combination of these fronts with a novel algorithm to obtain a $q$-sparse matrix.

This paper is organized as follows. 
In \cref{sec:relwork}, we present existing approaches for sparse MNNLS, and we detail the three algorithms used to generate Pareto fronts.
In \cref{sec:algo}, we present the main contribution of this work, that is, an algorithm to solve Problem \eqref{eq:matwisemnnls}.
We illustrate the effectiveness of our proposed method with experiments on real-world facial and hyperspectral image datasets \revise{and on synthetic datasets in \cref{sec:xp}.
We} conclude in \cref{sec:conclu}.

\section{Related work}
\label{sec:relwork}

Most approaches that tackle sparse MNNLS were actually introduced in the context of sparse NMF.
Since its very introduction by \cite{lee1999learning}, NMF is appreciated for the sparsity of the produced factors. 
A variety of works have been proposed to further enhance this sparsity, making \emph{sparse NMF} one of the most popular variants of NMF.
Many authors worked on the $\ell_1$-penalized formulation, notably \cite{hoyer2002non,eggert_sparse_2004,kim2007sparse,cichocki_flexible_2008,gillis2012sparse}.
This formulation uses the \lonenorm{} as a convex surrogate of the \lzeronorm{} to ease the computation, but it present\revise{s} several disadvantages, see \cref{subsec:homotopy} for a detailed explanation.

To avoid the issues linked to the $\ell_1$-penalty, \cite{hoyer2004non} introduced a more explicit sparsity measure based on the ratio between the  $\ell_1$-norm and the  $\ell_2$-norm, and he considered an NMF variant with a column-wise constraint on this measure.
Other works considered the \lzeronorm{} formulation, that can be decomposed into a series of $k$-sparse NNLS subproblems.
We can cite \cite{aharon2005ksvd,morup2008approximate,peharz_sparse_2012}.
\cite{cohen_nonnegative_2019-1} proposed a method that solves \emph{exactly} the $k$-sparse NNLS subproblems using a bruteforce approach.
\cite{nadisic2020exact} extended this work by replacing the bruteforce subroutine by a dedicated branch-and-bound algorithm.
To the best of our knowledge, no existing work considered a matrix-wise \lzero{} constraint.

Some similar models have been studied, such as simultaneous sparse approximation \citep{tropp2006algorithms,stojnic2009reconstruction}, where $X$ is constrained to be block-sparse, 
that is, to have sparse columns sharing the same support. 
The assumptions of these models, the algorithms to solve them, and their applications are far from our focus, so detailing them is out of the scope of this article.

In the following, we detail three types of sparse NNLS algorithms on which we will rely to generate Pareto fronts, that is, to solve Problem~\eqref{eq:biobj}.

\subsection{Arborescent}\label{subsec:arbo}

\begin{figure}[htb]
  \centering
  \includegraphics[width=0.8\textwidth]{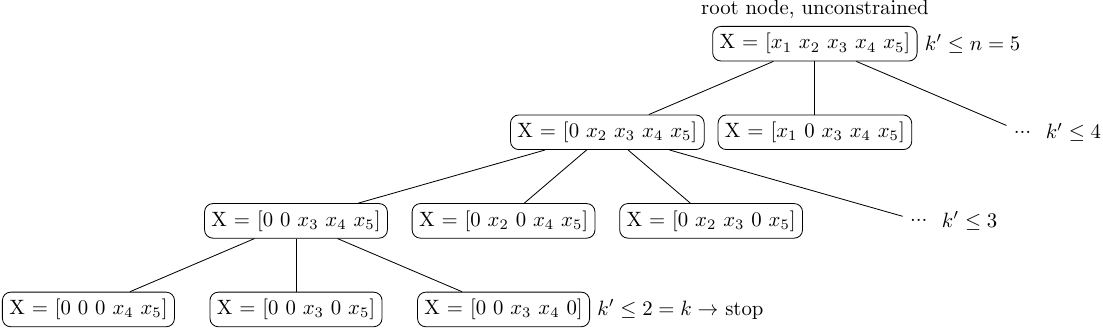}
  \caption{Example of the search tree explored by Arborescent, for $n=5$ and $k=2$.}
  \label{fig:arbo}
\end{figure}

The algorithm Arborescent \citep{nadisic2020exact} is a branch-and-bound algorithm designed to solve exactly $k$-sparse NNLS problems.
In a nutshell, Arborescent enumerates the possible supports (that is, the possible patterns of zeros) on a search tree, as shown \revise{in} \cref{fig:arbo}.
Every node represents an over-support $\mathcal{K}$, that is, the set of entries that are not constrained to 0.
Exploring a node means solving the NNLS subproblem
\begin{equation}
f^*(\mathcal{K}) = \min || A(:,\mathcal{K}) x(\mathcal{K}) - b ||_2^2 \text{ such that } {x(\mathcal{K}) \geq 0} \revise{,}
\end{equation}
\revise{where $x(\mathcal{K})$ denotes the subvector of $x$ with indices in the set $\mathcal{K}$}.
This subproblem can be solved with any standard NNLS solver, and here it is done with an \emph{active-set} algorithm \citep{portugal1994comparison}, which provides an exact solution in a finite number of iterations.
The value $f^*(\mathcal{K})$ is the \emph{error} associated with the node corresponding to $\mathcal{K}$.
To prune this tree, Arborescent uses the fact that in any optimization problem, when adding constraints, the solution cannot improve.
By doing a depth-first exploration, we can quickly find feasible solutions and then prune efficiently large parts of the search space.

An extension of this algorithm \citep{nadisic2021biobjective} computes exactly the Pareto front corresponding to the biobjective problem~\eqref{eq:biobj}.
It is based on the fact that, when computing the $k$-sparse solution to \revise{an} NNLS problem, Arborescent also computes all $k'$-sparse solutions for $k' \in \{k,\ldots,r\}$.
If we set $k=1$, then we compute the entire Pareto front.
If $k>1$, we compute only a portion of it.

This algorithm is fast in practice when the dimension $r$ is small, which is generally the case in hyperspectral unmixing.
However, it is still computationally expensive and quite slow for problems of large dimensions, when $r$ is larger than a few tens.
For this reason, practitioners often prefer other sparsity-inducing approaches, such as greedy algorithms or $\ell_1$-regularization.

Other works tackled the \lzero{}-constrained problem exactly, but without nonnegativity constraints, see for example \cite{ben2021global} and the references therein.
It may be possible to adapt them in the nonnegative setting, but this is still an open problem and out of the scope of this article.
To best of our knowledge, no other work considered computing the Pareto front of the biobjective sparse problem.

\subsection{Greedy algorithms}\label{subsec:greedy}

Greedy algorithms are one of the most popular approaches for solving Problem~\eqref{eq:ksnnls}.
They start with an empty support ($x_i = 0$ for all $i$), and select components one by one to enrich the support, until the target sparsity $k$ is reached.
The selection of a component is done greedily by choosing the component minimizing the residual error.
Orthogonal versions of these methods, such as \emph{Orthogonal Least Squares} (OLS)~\citep{chen1989orthogonal} and \emph{Orthogonal Matching Pursuit} (OMP)~\citep{pati1993orthogonal}  make sure a component can be selected only once.
Nonnegative variants have been recently proposed; see for example~\cite{nguyen2019nonnegative} and the references therein.
In general, these algorithms do not give the globally optimal solution.
Theoretical recovery guarantees exist, but they are restrictive~\citep{tropp2004greed,soussen2013joint}.

Interestingly, because they select components one after the other, greedy algorithms can be used as a proxy to compute an approximation of the Pareto front of the corresponding biobjective sparse NNLS problem.
Indeed, the solution at the $i$-th iteration of the algorithm is $i$-sparse.
By running the algorithm with a sparsity target $k$, we also compute, as a side effect, some $k'$-sparse solutions for $k' \in \{1,\ldots,k\}$.
Therefore, to obtain an approximation of the Pareto front, it suffices to return all intermediate solutions instead of only the final one.

In this paper, we will only focus on Nonnegative OMP (NNOMP) for conciseness, but this approach could be easily generalized to similar greedy algorithms.
NNOMP was first introduced by \cite{bruckstein2008uniqueness}, but many variants exist, and implementations details can have a significant impact on the performance of these algorithms.
Reviewing them is out of the scope of this article, and we refer the interested reader to \cite{nguyen2019nonnegative}.

\subsection{Homotopy}\label{subsec:homotopy}

The \lonenorm{}, defined as $\|x\|_1 = \sum_{i=1}^{r} | x_i |$, is a convex surrogate of the \lzeronorm{}, it is therefore easier to optimize while being able to promote sparsity.  
The $\ell_1$-regularization consists in penalizing the solution in the objective function of (\ref{eq:nnls}),  leading to the following problem, referred to as $\ell_1$-NNLS, 
\begin{equation}\label{eq:l1nnls}
    \min\limits_{x\in \mathbb{R}^r}  \| Ax - b \|_2^2 + \lambda \|x\|_1 \quad \text{ such that } \quad x \geq 0.
\end{equation}
Note that without the nonnegativity constraint, this is the well-known LASSO model~\citep{tibshirani1996regression}.
Problem~\eqref{eq:l1nnls} can be thought of as the weighted-sum form of a biobjective problem, where the objectives are minimizing the reconstruction error $ \| Ax - b \|_2^2$ on one hand, and minimizing the \lonenorm{} of the solution $\|x\|_1$ on the other hand.
The parameter $\lambda$ controls the trade-off between the two objectives.

Despite its popularity, this technique suffers from several drawbacks.
In particular, there is no explicit relation between the parameter $\lambda$ and the sparsity of the solution, hence choosing an appropriate value for $\lambda$ can be tricky, and often involves a tedious trial-and-error process.
Also, the $\ell_1$-penalty introduces a bias.
Although there exist theoretical guarantees for support recovery, such as the \emph{Exact Recovery Condition}~\citep{tropp2006just}, they are restrictive and often not realistic in practice.

To overcome these issues, \emph{homotopy} algorithms have been introduced.
They generate the full \emph{regularization path} of a given $\ell_1$-NNLS problem, that is, the set of the solutions for all possible values of $\lambda$.
They allow the user to choose the relevant solution within this path, each solution representing a different trade-off between sparsity and reconstruction error.
The first homotopy algorithm has been introduced by \cite{osborne2000new}, for sparse least squares with no nonnegativity constraint. 
Variants have been developed by \cite{efron2004least} and \cite{donoho2008fast}.
\cite{kim2013regularization} introduced a variant to deal specifically with the nonnegativity constraint.

In a nutshell, the homotopy algorithm uses the KKT conditions (necessary conditions for optimality) to first find  the value $\lambda_{\max}$ for which the optimal solution of $\ell_1$-NNLS is $x=0$ for any $\lambda \geq \lambda_{\max}$, and then to compute the next smaller values of $\lambda$ for which the support (that is, the set of non-zero entries) of the optimal solution changes (one zero entry becomes non-zero, or the other way around).
This is similar in spirit to active-set methods, akin to the simplex algorithm for linear programming.
These values of $\lambda$ are called \emph{breakpoints}, between which the support of the optimal solution does not change;
see \cref{fig:path,fig:pareto} for an illustration.
\begin{figure}[ht]
  \begin{minipage}{0.45\linewidth}
    \centering
    \definecolor{mypurple}{rgb}{0.5961,0.3059,0.6392}\definecolor{mygreen}{rgb}{0.7020,0.8706,0.4118}\definecolor{myorange}{rgb}{0.9843,0.5020,0.4471}

\begin{tikzpicture}[scale=0.5]

\def\bpa{8}
\def\bpb{6}
\def\bpc{3.5}
\def\bpd{2}

\draw[->] (0,0) -- (9,0) node[anchor=north] {$\lambda$};
\draw[->] (0,0) node[anchor=east] {$x_i = 0$} -- (0,4) node[anchor=east] {$x_i$};

\draw	(0,0) node[anchor=north] {$\lambda = 0$}
        (\bpd,0) node[anchor=north, name=lambdad] {$\lambda_4$}
		(\bpc,0) node[anchor=north, name=lambdac] {$\lambda_3$}
		(\bpb,0) node[anchor=north, name=lambdab] {$\lambda_2$}
		(\bpa,0) node[anchor=north, name=lambdaa] {$\lambda_{max}$};

\draw[->, bend left=15] (lambdaa) edge (lambdab) (lambdab) edge (lambdac) (lambdac) edge (lambdad);
		
\draw[dotted] (\bpd,0) -- (\bpd,4)
              (\bpc,0) -- (\bpc,4)
              (\bpb,0) -- (\bpb,4)
              (\bpa,0) -- (\bpa,4);
		
\draw[very thick,myorange] (0,3.5) node[anchor=east] {$x_1$}
                      (0,3.5) -- (\bpd,2.5)
                      (\bpd,2.5) -- (\bpc,3)
                      (\bpc,3) -- (\bpb,2)
                      (\bpb,2) -- (\bpa,0)
                      (\bpa,0) circle (0.1);
                      
\draw[very thick,mygreen] (0,0) node[anchor=south west] {$x_2$}
                     (0,0) -- (\bpd,0)
                     (\bpd,0) -- (\bpc,2)
                     (\bpc,2) -- (\bpb,0)
                     (\bpb,0) -- (\bpa,0)
                     (\bpb,0) circle (0.1)
                     (\bpd,0) circle (0.1);
                      
\draw[very thick,mypurple] (0,2) node[anchor=east] {$x_3$}
                      (0,2) -- (\bpd,1.5)
                      (\bpd,1.5) -- (\bpc,0)
                      (\bpc,0) -- (\bpb,0)
                      (\bpb,0) -- (\bpa,0)
                      (\bpc,0) circle (0.1);
                      
\end{tikzpicture}
     \caption{Example of a solution path of a homotopy algorithm, depending on $\lambda$, for a problem of 3 variables. Vertical dotted lines correspond to breakpoints.}
    \label{fig:path}
  \end{minipage}
  \hfill
  \begin{minipage}{0.5\linewidth}
    \centering
    \definecolor{mypurple}{rgb}{0.5961,0.3059,0.6392}\definecolor{mygreen}{rgb}{0.7020,0.8706,0.4118}\definecolor{myorange}{rgb}{0.9843,0.5020,0.4471}

\begin{tikzpicture}[scale=0.5]

\coordinate (A) at (5,0);
\coordinate (B) at (3.7,0.5);
\coordinate (C) at (2.5,1.2);
\coordinate (D) at (1.5,2.5);
\coordinate (E) at (1,3.5);

\draw[->] (0,0) -- (7,0) node[anchor=west] {$\| Ax-b \|_2^2$};
\draw[->] (0,0) node[anchor=east] {0} -- (0,4) node[anchor=east] {$\| x \|_1$};

\draw[thick] (A) -- (B)
             (B) -- (C)
             (C) -- (D)
             (D) -- (E);

\filldraw[thick,draw=myorange,fill=myorange!50] (A) circle (0.1) node[anchor=south west,myorange] {$\lambda_{max}$};\draw (A) node[below]{$\|b\|_2^2$};
\filldraw[thick,draw=mygreen,fill=mygreen!50] (B) circle (0.1) node[anchor=south west,mygreen] {$\lambda_2$};
\filldraw[thick,draw=mypurple,fill=mypurple!50] (C) circle (0.1) node[anchor=south west,mypurple] {$\lambda_3$};
\filldraw[thick,draw=mygreen,fill=mygreen!50] (D) circle (0.1) node[anchor=south west,mygreen] {$\lambda_4$};
\filldraw[thick,draw=black,fill=black!50] (E) circle (0.1) node[anchor=south west] {$\lambda = 0$};

\end{tikzpicture}     \caption{Trade-off between the $\ell_1$-norm and the error $\|Ax-b\|_2^2$ corresponding to the solution path in \cref{fig:path}.}
    \label{fig:pareto}
  \end{minipage}
\end{figure}
Although the nonnegative homotopy algorithm is not an original contribution, we include in \cref{app:homotopy} its detailed description and justification.

The strength of the homotopy algorithm is to generate the full \emph{regularization path}, that is, the set of optimal solution for all possible values of $\lambda$, for the same cost as a standard active-set algorithm.
The solutions in this path represent different tradeoffs between reconstruction error and sparsity, and as such this path can be seen as an approximation of the Pareto front in Problem \eqref{eq:biobj}.

\section{Solving matrix-wise q-sparse MNNLS}
\label{sec:algo}

In this section, we study how to tackle matrix-wise $q$-sparse MNNLS, that is, Problem \eqref{eq:matwisemnnls}.
First, we present the key contribution of this work, that is a two-step algorithm to solve Problem \eqref{eq:matwisemnnls}.
Then, we show how the greedy algorithm NNOMP can be implemented specifically for Problem \eqref{eq:matwisemnnls} to avoid the costly reformulation from \cref{eq:mnnlsvec}.

\subsection{Our key contribution: a two-step algorithm for matrix-wise $q$-sparse MNNLS}\label{subsec:ouralgo}

In this section we present the main contribution of this paper, that is, a two-step algorithm to solve Problem \eqref{eq:matwisemnnls}.
This algorithm is called Salmon\footnote{The name stands for SALMON Applies \lzero{}-constraints Matrix-wise On NNLS problems.} and it is detailed in \cref{alg:salmon}.
The motivation behind it is to divide the large matrix-wise problem into small one-column subproblems, solve them, and then combine their solutions to build a global solution.

\begin{algorithm}[htp]

\SetKwFor{ForEach}{for each}{}{}

  \KwIn{Data matrix $B \in \mathbb{R}^{m \times n}$, dictionary $A \in \mathbb{R}^{m \times r}$, sparsity target $q \in \mathbb{N}$}
  \KwOut{$X^* \in \mathbb{R}^{r \times n}$ }

  \DontPrintSemicolon
  
  Init matrix $C \in \mathbb{R}^{r \times n}$,
  for all $i$ and $j$, $C(i,j) \gets \infty$ \label{sal:initc}\\
  Init solutions $Sol_{i,j}$ for all $i$ and $j$

  \ForEach{ $j \in \{ 1 , \ldots , n \}$ \label{sal:outerloop}}
  {         
      $\{x^*_t\}_{\forall t} \gets$ front\_generator($A$,$B(:,j)$) \label{sal:callhomotopy}  \tcc{Arborescent, NNOMP, or homotopy}
      \ForEach{ $t$ \label{sal:loopbuildc}}
      {
          $k \gets \| x^*_t \|_0 $  \label{sal:findk}\\
          $err \gets \| B(:,j) - A x^*_t \|_2^2$ \label{sal:err}\\
        \ForEach{$i \in \{k,\ldots,r\}$ \label{sal:foriinkr}}
        {
            \If{$err < C(i,j)$ \label{sal:iferrless}}
            {
                $C(i,j) \gets err $  \label{sal:updtckj}\\
                $Sol_{i,j} \gets x^*_t$ \label{sal:updtsol}
            }
        }
          
      }
  }

  Init cursors, for all $ j \in \{1,\ldots,n\}, k_j \gets 0$ \label{sal:initcursor}\\
  Init matrix $\mathcal{G} \in \mathbb{R}^{r \times n}$\\
  
  \ForEach{$i \in \{1,\ldots,r\} , j \in \{1,\ldots,n\}$ \label{sal:foreachg}}
  {
    $\mathcal{G} (i,j) \gets ( C(0,j) - C(i,j) ) / i$ \label{sal:buildg}
  }
  
  \While{$ \sum_j k_j < q$ \textbf{\textup{and}} $\max_{(i,j)} \mathcal{G}(i,j) > 0$}
  {
    
    $i^*, j^* \gets \argmax_{(i,j)} \mathcal{G}(i,j)$ \label{sal:findmax} \\
    $\delta \gets i^* - k_{j^*}$\label{sal:delta} \\
    $k_{j^*} \gets i^*$ \label{sal:updtcur} \\
    
    \ForEach{$i \in \{1,\ldots,i^*\}$ \label{sal:beginupdtg}}
    {
        $ \mathcal{G}(i,j^*) \gets 0 $ \label{sal:updtgzero}\\
    }
    \ForEach{$i \in \{i^* + 1, \ldots, r\}$}
    {
         $ \mathcal{G}(i,j^*) \gets \frac{C(i^*,j^*) - C(i,j^*)}{i-i^*} $ \label{sal:endupdtg} \\
    }

    \If{$ \sum_j k_j > q - r + 1 $ \label{sal:ifnearend}}
    {
        \ForEach{$(i,j)$ \textnormal{such that} $i \in \{ k_j + 1 , \ldots , \min(r, k_j + q - \sum_{l > j} k_l) \} $ \label{sal:forupdtgzero2}}
        {
            $ \mathcal{G}(i,j) \gets 0 $ \label{sal:updtgzero2}\\
        }
    }
    
  }
  
\ForEach{$j \in \{1,\ldots,n\}$ \label{sal:loopbuildh}}
  {
    $X^*(:,j) \gets Sol_{k_j, j}$ \label{sal:end}
  }
  
  \caption{The algorithm Salmon}
  \label{alg:salmon}
\end{algorithm}

Step 1 corresponds to lines \ref{sal:initc} to \ref{sal:updtsol}.
It consists in, given the data matrix $B$ and the dictionary $A$, running an algorithm to generate a Pareto front for every column of $B$, that is, with input $A$ and $b = B(:j)$ for all $j$.
This can be done by any of the three Pareto-front-generating methods presented in \cref{sec:relwork}; exactly with Arborescent and approximately with NNOMP and the homotopy algorithm.
From these fronts, we build a cost matrix $C$ where each column represents a column $j$ of $X$, each row represents a $k$-sparsity between $0$ and $r$, and each entry is the reconstruction error of the $k$-sparse solution of column $j$.
Formally,  for all $i \in \{0,\ldots,r\}$ and $j \in \{1,\ldots,n\}$,
\begin{equation*}
    C(i,j) \quad \approx \quad 
    \min_{x \geq 0} \| B(:,j) - A x \|_2^2 \text{ s.t. } \|x\|_0 \leq i ,
\end{equation*}
and $Sol_{i,j}$ stores the corresponding argmin, that is the best $i$-sparse solution for column $j$.

\begin{remark}
The algorithms used to generate the Pareto front in step 1 do not necessarily generate one $k$-sparse solution for a each $k$; they may generate more than one solution, or no solution at all for a given $k$. 
For example, NNOMP selects columns of $A$ sequentially, but some of them may not have a positive weight when solving the corresponding NNLS. 
The loop on \cref{sal:foriinkr} ensures that, if there exists some $k$ for which no $k$-sparse solution is generated, then the $(k-1)$-sparse solution is used instead (or the $(k-2)$-sparse one if no $(k-1)$-sparse solution is generated, and so on).
The condition on \cref{sal:iferrless} ensures that, if there are several $k$-sparse solutions for a given $k$, then only the best one is kept.
\end{remark} 

Once $C$ is computed, step 2 consists in selecting one solution per column to build the solution matrix $X$, that is, it consists in choosing the sparsity level for each column of $X$.
This selection step is a combinatorial problem, similar to an assignment problem. 
Let us define the binary variables $z_{i,j} \in \{0,1\}$ for $i \in \{0,1,\ldots r\}$ and $j \in \{1,2,\ldots n\}$ such that $z_{i,j} = 1$ if and only if the $j$th column of $X$ is $i$-sparse.
\revise{Note that $i$ can be equal to 0, corresponding the the zero vector.}
The variable $z$ encodes which sparsity level is selected for each column of $X$. 
Given the cost matrix $C$ computed in step 1, 
step 2 requires to solve the following integer program
\begin{align} 
    \min_{z \in\{0,1\}^{r \times n}} & \sum_{i,j} z_{i,j} C(i,j) \nonumber  \\ 
    \text{ such that }
    & \sum_i z_{i,j} = 1 \text{ for all $j$},  
    \; \text{ and } \; 
    \sum_{i,j} i \, z_{i,j} \leq q. \label{eq:step2} 
\end{align} 
The objective is to minimize the reconstruction error, while the first  constraints impose that each column of $X$ has a single sparsity level, and the second that the total number of non-zero entries of $X$ does not exceed $q$.

We propose a greedy selection algorithm to solve~\eqref{eq:step2}, see lines \ref{sal:initcursor} to \ref{sal:end} of \cref{alg:salmon}. As we will prove in Theorem~\ref{th:nearopti}, this greedy strategy is nearly optimal. It works as follows. 
For each column $j$, the scalar $k_j$ indicates its sparsity level at the current iteration, that is, at every iteration, the $j$th column of $X$ is $k_j$-sparse.  
We initialize the algorithm with the 0-sparse solution (the vector of all zeros) for each column (\cref{sal:initcursor}), that is, $k_j = 0$ for all $j$.  
Note that $\sum_j k_j$ corresponds to the current sparsity level of the solution.

At each iteration, we will decrease the sparsity of a single column of $X$. To pick that column in an optimal way, let us define the matrix $\mathcal{G}$, with the same dimensions as $C$, as follows: 
at any iteration, the entry $\mathcal{G}(i,j)$ is equal to the potential gain in reconstruction error if the $j$th column of $X$ goes from $k_j$-sparse to $i$-sparse divided by the sparsity difference between these two solutions (that is, $i-k_j$). 
In particular, at the first iteration (\cref{sal:buildg}), 
when $X = 0$, we have 
\[ 
\mathcal{G}(i,j) = \frac{C(0,j) - C(i,j)}{i} 
\quad \text{ for all } i,j .
\] 
Let us denote by $(i^*, j^*)$ the position of the largest entry of  $\mathcal{G}$. 
Given a current solution, the column that will decrease the error $\| B-AX \|_F^2$ the most by decreasing its sparsity is the one corresponding to the column of $\mathcal{G}$ with the largest entry, that is, the $j^*$th column. 
Finding this entry is cheap in practice, as we update a sorted list of the maximum entry of each column of $\mathcal{G}$.
We denote the quantity $\delta$ as the difference in sparsity of the selected column before and after it has been updated, 
that is, $\delta = i^* - k_{j^*}$.  
After the value of $k_{j^*}$ has been updated to $i^*$, we update  the entries of the $j^*$th column of $\mathcal{G}$ accordingly  (\cref{sal:endupdtg}). Note that $\mathcal{G}(i,j^*) = 0$ for all $i \leq i^*$. 

To avoid generating a too dense solution (recall $\sum_j k_j$ must be smaller than $q$), the procedure on  \cref{sal:ifnearend,sal:forupdtgzero2,sal:updtgzero2} sets to zero the entries of $\mathcal{G}$ whose selection would lead to a total sparsity $\sum_j k_j$ larger than $q$. 
Note that thanks to the condition on \cref{sal:ifnearend}, this procedure is executed only for the last few iterations of Salmon.
Indeed, the maximum increase $\delta$ is $r$, so this procedure would not be useful as long as $\sum_j k_j \leq q-r$.

When the sum of the sparsity levels equals $q$, or when all entries of $\mathcal{G}$ are zero, we stop the procedure and we build the final $q$-sparse solution $X$ by selecting for each column the solution corresponding to its final sparsity level (\cref{sal:loopbuildh}).

Although the selection algorithm of step 2 is greedy, since we perform an optimal selection at each iteration, it is able to generate a near-optimal solution to Problem \eqref{eq:step2}, as shown below. 
\begin{theorem}[Near-optimality of the selection step]\label{th:nearopti}
Given that $C$ is non-increasing by columns, that is, $C(i,j) \leq C(i',j)$ for all $i' \leq i$ and all $j$, the proposed selection step of Salmon (lines \ref{sal:initcursor} to \ref{sal:end} of \cref{alg:salmon}) computes a near-optimal solution of \eqref{eq:step2} in the following sense. 
Denoting  
\begin{itemize}
    \item[$\bullet$] $f(z)$$=$$\sum_{i,j} z_{i,j} C(i,j)$ the value of the objective function of~\eqref{eq:step2} for a solution~$z$, 
    \item[$\bullet$] $z^*$ an optimal solution to Problem \eqref{eq:step2}, and 
    \item[$\bullet$] $z_{Sal}$ the solution computed by the selection step of Salmon, 
\end{itemize} 
we have
\begin{equation*}
    f(z^*) \leq f(z_{Sal}) \leq f(z^*) + \max_j \| C(:,j) \|_\infty .
\end{equation*}
\end{theorem}
\begin{proof}
First, note that our proposed greedy algorithm generates a feasible solution of~\eqref{eq:step2} since we make sure that  
$\sum_j k_j$ remains smaller than $q$, and hence, 
by optimality of
$z^*$, $f(z^*) \leq f(z_{Sal})$. 

Let us now show that $f(z_{Sal}) \leq f(z^*) + \max_j \| C(:,j) \|_\infty$. 
Our greedy procedure is similar to a dynamic programming approach. 
In fact, let us denote the optimal objective function value of \eqref{eq:step2} as $f^*(q)$ that depends on the parameter $q$, the global sparsity level allowed; note that $f(z^*) = f^*(q)$. 
The greedy algorithm is initialized with  
$z_{0,j} = 1$ for all $j$, that is, $X = 0$, which is optimal for $q=0$ (it is the only feasible solution), and hence gives $f^*(0) = \sum_j C(0,j)$.  
It then progressively decreases the sparsity to reduce the objective the most at each iteration.   
At each iteration of the greedy algorithm, the support of a single column of $X$ is increased in order to maximize the ratio between the decrease in objective function value and the decrease of sparsity.  
Since the columns of $X$ do not interact with each other in the objective function and since $C$ is non-increasing in each column, 
the greedy solution cannot possibly be improved as long as $\delta \leq q - \sum_j k_j$, that is, as long as this optimal way of picking a column is allowed by the global sparsity level. In summary, our greedy algorithm produces intermediate optimal solutions of~\eqref{eq:step2} with objective $f^*(\sum_j k_j + \delta)$ as long as $\sum_j k_j + \delta \leq q$.     

The only moment when the greedy algorithm might fall short of global optimality is during the last iterations: 
if at some point the optimal move is to increase the support of a column in such a way that the total sparsity would exceed $q$ (that is, $\sum_j k_j + \delta > q$), then our greedy algorithm may not be optimal because, to allow that move, we might need to reduce the support of another column, which the greedy approach does not allow. 
Making that move anyway would generate an optimal solution with global sparsity $q' = \sum_j k_j + \delta > q$ which would not be feasible for~\eqref{eq:step2}. 
Observe that  
\begin{itemize}
    \item[$\bullet$] 
    $f(z_{Sal}) \leq f^*(q'- \delta)$ since the greedy algorithm keeps improving the $(q'-\delta)$-sparse solution in its next iterations (although possibly not optimally). 

    \item[$\bullet$] 
    $f^*(q') \geq f^*(q'- \delta) - \max_j \|C(:,j)\|_\infty$ since the move from $q' - \delta$ to $q'$ using the greedy strategy is optimal and, in the worst case, will decrease the sparsity level of a column from $r$ to $0$ reducing the error by at most $\max_j \|C(:,j)\|_\infty$. 
    
    \item[$\bullet$] $f^*(q') \leq f^*(q) \leq f^*(q'-\delta)$ 
    since $q'-\delta \leq q \leq q'$. 
\end{itemize}
Combining these observations, we obtain 
\[
f^*(q) = f(z^*) 
\geq 
f^*(q') 
\geq 
f^*(q' - \delta) - \max_j \|C(:,j)\|_\infty
\geq 
f(z_{Sal}) - \max_j \|C(:,j)\|_\infty , 
\] 
which gives the result. \eproof 
\end{proof}

Note that, in most practical cases, such as hyperspectral unmixing, we have $n \gg r$ and hence $\max_j \| C(:,j) \|_\infty$ is negligible compared to $f(z^*)$.
In addition, it is rather unlikely that the greedy algorithm needs to increase the sparsity level of one column from 0 to $r$ at the last step. In fact, in our experiments, we observed that $\delta$ is in most cases equal to $1$.  
When $\delta$ is equal to $1$ in the last $r$ steps, the greedy algorithm is globally optimal (or, more generally, when $\delta = q - \sum_k k_j$ in the last step).
For example, for the datasets used in the numerical experiments (\cref{sec:xp}) and with the three sparse NNLS algorithms to generate the Pareto fronts (that is, the matrix $C$), the greedy selection algorithm generated a guaranteed globally optimal solution (that is, $\delta = q - \sum_k k_j$ in the last step) in 19 out of 22 cases.

In practice, the global optimality of Salmon to solve the sparse MNNLS problem~\eqref{eq:matwisemnnls} therefore heavily relies on step 1. 
If step 1 is done with Arborescent, then the Pareto fronts are computed optimally and Salmon computes a near-optimal solution to Problem \eqref{eq:matwisemnnls}.   
Otherwise, it only computes an approximate solution, although there exist some conditions under which NNOMP or the homotopy algorithm do recover the true Pareto fronts, see \cref{subsec:greedy,subsec:homotopy}.

\paragraph{Computational cost of Salmon}  

The cost of step 1 depends on the algorithm used to generate the Pareto fronts.
In all cases, the $n$ biobjective subproblems are solved independently, so the cost of step 1 grows linearly with $n$.
For one subproblem, that is, to generate one Pareto front, we have that 
\begin{itemize}
    \item The cost of Arborescent depends on the number of nodes explored in the branch and bound. 
    In the worst case, it is of the same order as the bruteforce algorithm, and requires to solve $\binom{r}{k}$ NNLS subproblems, while, in the best case, it is of the order of $r$~\citep{nadisic2020exact}.
    Empirically, the cost is far from the worst case but grows faster than linear with $r$. 
    
    \item The cost of NNOMP is in $\mathcal{O}(mr^4)$ operations~\citep{yaghoobi2015fast}.
    
    \item The cost of the homotopy algorithm is of the same order as an active-set algorithm and requires at least $\mathcal{O}(r^4)$ operations, see \cref{app:homotopy}.
\end{itemize}

Given $C$, the selection step consists in building $\mathcal{G}$ in $\mathcal{O}(r n)$ operations, 
then iterating $q$ times (in the worst case) to select a solution.
As we maintain updated a sorted list to avoid recomputing the maximum at each iteration, this is done in $\mathcal{O}(q \log(n))$ operations.
Since $q$ is in the order of $rn$ in the worst case, 
the cost of the selection step is dominated by the cost of the Pareto-front-generating step.

\subsection{Adapting NNOMP for matrix-wise $q$-sparse MNNLS}\label{subsec:nnompg}

Let us now adapt NNOMP for the matrix-wise $q$-sparse MNNLS problem.
To avoid the overcost of solving the vectorized problem \eqref{eq:mnnlsvec} directly with NNOMP, we can adapt NNOMP to handle the matrix form of Problem \eqref{eq:matwisemnnls}.
We detail the adaptation in \cref{alg:nnompg}; note that it is not an original contribution and seems to be common knowledge in the sparse approximation community.
The solution produced by this algorithm is strictly equivalent to the one produced by direct solving of \eqref{eq:mnnlsvec}.
Our goal in introducing it is to show how using NNOMP within the two-step algorithm Salmon is advantageous compared to using NNOMP directly on Problem \eqref{eq:matwisemnnls}. \begin{algorithm}[htp]

  \KwIn{$B \in \mathbb{R}^{m \times n}$, $A \in \mathbb{R}^{m \times r}$, $q \in \mathbb{N}$}
  \KwOut{$X$, approximate solution to Problem \eqref{eq:matwisemnnls}}

  \DontPrintSemicolon
  
  $X \gets 0^{r \times n}$ ; $S \gets \emptyset $ ; $R_S \gets B$ \label{nnomp:init}
  
  \While{ $|S| < q$ and $\max_{(i,j)} (A^\top R_S)_{i,j} > 0$ \label{nnomp:while}}
  {
    $(i^*, j^*) \gets \argmax_{(i,j) \notin S} A^\top R_S$ \label{nnomp:select}\\
    $S \gets S \cup (i^*, j^*)$ \label{nnomp:updateS}\\
    $X \gets \min_X \| A_S - B_S X_S \|_F^2$ \label{nnomp:updateX}\\
    $R_S \gets B-AX$ \label{nnomp:updateRs}\\
    $S \gets \{ (i,j) : X(i,j) > 0 \} $  \label{nnomp:compress}
  }
  
  \caption{The greedy algorithm NNOMP adapted for matrix-wise $q$-sparse MNNLS.}
  \label{alg:nnompg}
\end{algorithm} 
On \cref{nnomp:init}, we initialize $X$ as a zero matrix, $S$ as an empty support, and the residual $R_S$ as equal to $B$.
Then, we select greedily entries to add to the support, while the cardinality of the support is lower than $q$ and the residual greater than zero (\cref{nnomp:while}).
The greedy selection on \cref{nnomp:select} consists in choosing the entry that maximizes the decrease of the residual error; this entry is added to the support on \cref{nnomp:updateS}.
Then, we update $X$ on \cref{nnomp:updateX} (this is an NNLS problem, that we solve with an active-set algorithm), and the residual $R_S$ on \cref{nnomp:updateRs}; note that in practice we only need to update the $j^*$-th column of $X$ and $R_S$.
On \cref{nnomp:compress} we perform a support compression, that is, we restrain the support to the non-zero entries of $X$. 
This last step is necessary because of the nonnegativity constraint, that may put to zero an entry that was selected at a previous iteration. 

Other greedy algorithms could be adapted similarly, but here we focus only on NNOMP for conciseness.
Also, our goal is to study how the original NNOMP compares to NNOMP used within the two-step approach Salmon, rather than comparing different greedy algorithms with each other.
The homotopy algorithm may also be similarly adaptable, but this is not trivial and out of the scope of this article.

\section{Experiments}
\label{sec:xp}

In this section, we study the performance of the proposed algorithm Salmon on the unmixing of 7 datasets: 3 faces datasets and 4 hyperspectral images.
\revise{Then, we study the evolution of the computing time of the algorithm when the sparsity parameter $q$ varies.
Finally, we test on synthetic datasets the ability of Salmon to recover the underlying solution in the presence of noise, when the sparsity varies between columns, with both well-conditioned and ill-conditioned data.
}

\subsection{Data}

In the faces datasets, each column of $B$ corresponds to a pixel and each row to an image (that is, $B(i,j)$ is the intensity of pixel $j$ in image $i$).
It is well-known that NMF will extract facial features as the rows of matrix $X$ \citep{lee1999learning}. 
As no groundtruth is available, we first compute $A$ with SNPA \citep{gillis2014successive}, an algorithm for separable NMF, setting the factorization rank $r$ as in the literature.
We then compute $X$ with our sparsity-enhancing method.
Imposing sparsity on $X$ means that we require that only a few pixels are contained in each facial feature \citep{hoyer2004non}. 
We consider the 3 widely used face datasets 
CBCL\footnote{Downloaded from \url{http://poggio-lab.mit.edu/codedatasets}},
Frey\footnote{Downloaded from \url{https://cs.nyu.edu/~roweis/data.html}}, and
Kuls\footnote{Downloaded from \url{http://www.robots.ox.ac.uk/}}.

Similarly 
a hyperspectral image is an image-by-pixel matrix where each image corresponds to a different wavelength. 
The columns of $A$ represent the spectral signature of the pure materials (also called endmembers) present in the image~\citep{bioucas2012hyperspectral}, and we use the ground truth $A$ from~\cite{zhu2017hyperspectral}. 
We compute $X$, whose columns represent the abundance of materials in each pixel.
It makes sense to impose $X$ to  be sparse as most pixels contain only a few endmembers~\citep{ma2013signal}.
We consider the 4 widely used datasets\footnote{Downloaded from \url{http://lesun.weebly.com/hyperspectral-data-set.html}} Jasper, Samson, Urban, and Cuprite.
The characteristics of these datasets are summarized in \cref{tab:datasets}.

\begin{table}[ht]
\caption{Summary of the datasets, for which $B \in \mathbb{R}^{m\times n}$ and $A \in \mathbb{R}^{m \times r}$.}\label{tab:datasets}
\centering \begin{tabular}{l|l|l|l|l}
Dataset & Type          & $m$  & $n$                      & $r$   \\ \hline
CBCL    & Faces         & 2429 & $19 \times 19 = 361$     & 49 \\
Frey    & Faces         & 1965 & $20 \times 28 = 560$     & 36 \\
Kuls    & Faces         & 20   & $64 \times 64 = 4096$    & 5  \\
Jasper  & Hyperspectral & 198  & $100 \times 100 = 10000$ & 4  \\
Samson  & Hyperspectral & 156  & $95 \times 95 = 9025$    & 3  \\
Urban   & Hyperspectral & 162  & $307 \times 307 = 94249$ & 6  \\
Cuprite & Hyperspectral & 188  & $250 \times 191 = 47750$ & 12
\end{tabular}
\end{table}

\subsection{Methods}

All methods have been implemented in Julia and run on a computer with a processor Intel Core i5-2520M @2.50GHz.
The code and experiment scripts are provided in an online repository\footnote{\url{https://gitlab.com/nnadisic/giant.jl}}.

We compare the following methods:
\begin{itemize}
    \item AS denotes an active-set algorithm that solves exactly the NNLS problem without sparsity constraint. It serves as a baseline to compare with the sparsity-constrained methods.
    
    \item \lone{}-CD denotes a coordinate descent algorithm with an \lone{} penalty. The penalty parameter $\lambda$ is fixed and the same for the whole matrix, as in \cite{kim2007sparse}. It has been tuned manually to reach the target sparsity. We compute the unbiased solution by running an active-set NNLS algorithm restrained to the non-zero elements of the \lone{}-penalized solution.
    
    \item Hcw denotes the homotopy algorithm described in \cref{subsec:homotopy}, that solves the column-wise $k$-sparse problem, as defined in~\eqref{eq:ksmnnls}\revise{.}
    For each column, we generate the regularization path, take the best $k$-sparse solution, and unbias it as described above.
    
    \item H+S corresponds to the two-step algorithm Salmon using the homotopy algorithm in step 1 (solutions are unbiased as above).
    
    \item OGcw stands for orthogonal greedy and denotes the NNOMP algorithm described in \cref{subsec:greedy}, that solves the column-wise $k$-sparse problem.
  
    \item OGg denotes the matrix-wise variant of NNOMP described in \cref{subsec:nnompg}.
   
    \item OG+S corresponds to the two-step algorithm Salmon using NNOMP in step~1. We generate the whole Pareto fronts, hence the column-wise sparsity target for NNOMP is $k=r$.
  
    \item ARBOcw denotes the branch-and-bound algorithm Arborescent described in \cref{subsec:arbo}, that solves the column-wise $k$-sparse problem.
  
    \item ARBO+S corresponds to the two-step algorithm Salmon using Arborescent in step 1. We generate the whole Pareto fronts, hence the column-wise sparsity target for Arborescent is $k=1$.
\end{itemize}

\revise{\subsection{Experiment 1: hyperspectral unmixing}}

\revise{In this experiment, we compare the performance of different variants of Salmon with the corresponding column-wise algorithms and with OGg.
For each dataset,} we choose the parameter $k$ by trial-and-error.
Unless stated otherwise, we define the sparsity parameter of matrix-wise methods as $q=k \times n$, which is equivalent to an average column-wise $k$-sparsity constraint. 

For every dataset, we run the nine methods, and measure the average column-sparsity of the given solutions (defined as the number of non-zero entries divided by the number of columns), the relative reconstruction error $\frac{\|B-AX\|_F}{\|B\|_F}$, and the computing time, for which we measure the median over 10 runs.
We set a timeout of 6000 seconds.
Note that, for a given dataset and with the same parameters, a given algorithm always gives the same output.
For the 3 methods based on NNOMP, we normalize the columns of the matrix $A$ before the computation.

The results of the experiments are summarized in \cref{tab:my-table}.
\begin{table}[ht]
\centering
\caption{
Results of the experiments, for the unmixing of facial and hyperspectral datasets.
Time is in seconds, relative error in percent, and sparsity is the average number of non-zero entries per column.
Numbers in bold represent, for a given setting, the error of ARBO+S and the best error among the other sparse methods.
For the variants of Salmon, a star $^*$ indicates that the greedy selection (step 2 of Salmon) is optimal (which can be checked easily: it requires $\delta = q - \sum_k k_j$ at the last iteration). 
Jasper is processed once with all algorithms for $k=q/n=2$, and once with matrix-wise algorithms for $q/n=1.8$ (which is not possible with column-wise algorithms).
}
\label{tab:my-table}
\resizebox{\textwidth}{!}{\begin{tabular}{ll|lllllllll}
        &          & AS    & \lone{}-CD & Hcw   & \textbf{H+S}    & OGcw & OGg  & \textbf{OG+S}  & ARBOcw  & \textbf{ARBO+S}   \\ \hline
CBCL    & Time     & 0.2   & 0.1   & 0.71  & 0.81  & 0.08  & 0.31  & 3.7   & timeout & timeout \\
$r=49$  & Error    & 12.04 & 17.37 & 16.19 & 13.22$^*$ & 13.12 & 12.35 & \textbf{12.3}$^*$  & -       & -       \\
$k=3$   & Sparsity & 6.64  & 3     & 2.69  & 3     & 2.37  & 3     & 3     & -       & -       \\ \hline
Frey    & Time     & 0.22  & 0.08  & 1.12  & 1.27  & 0.18  & 0.61  & 3.97  & timeout & timeout \\
$r=36$  & Error    & 19.35 & 21.76 & 23.22 & 20.75$^*$ & 21.35 & 19.86 & \textbf{19.8}  & -       & -       \\
$k=6$   & Sparsity & 12.29 & 6     & 5.52  & 6     & 4.64  & 6     & 6     & -       & -       \\ \hline
Kuls    & Time     & 0.17  & 0.12  & 0.18  & 0.17  & 0.28  & 1.82  & 0.5   & 0.67    & 1.41    \\
$r=5$   & Error    & 19.05 & 19.61 & 20.13 & 19.12$^*$ & 19.46 & \textbf{19.06} & \textbf{19.06}$^*$ & 19.42   & \textbf{19.06}$^*$   \\
$k=3$   & Sparsity & 3.45  & 3     & 2.86  & 2.99  & 2.7   & 3     & 3     & 2.76    & 3       \\ \hline
Jasper  & Time     & 0.34  & 0.22  & 0.38  & 0.48  & 0.39  & 6.08  & 1.12  & 1.21    & 1.93    \\
$r=4$   & Error    & 5.71  & \textbf{5.72}  & 6.99  & \textbf{5.72}$^*$  & 7.49  & 5.76  & 5.73  & 6.18    & \textbf{5.71}$^*$   \\
$k=2$   & Sparsity & 2.27  & 2     & 1.78  & 1.99  & 1.72  & 2     & 2     & 1.78    & 2       \\ \hline
Jasper  & Time     & -     & 0.18  & -     & 0.44  & -     & 5.26  & 1.15  & -       & 1.7     \\
$r=4$   & Error    & -     & 7.87  & -     & 5.95$^*$  & -     & 6.06  & \textbf{5.77}$^*$  & -       & \textbf{5.74}$^*$    \\
$q/n=1.8$ & Sparsity & -   & 1.8   & -     & 1.79  & -     & 1.8   & 1.8   & -       & 1.8     \\ \hline
Samson  & Time     & 0.22  & 0.24  & 0.2   & 0.26  & 0.31 & 3.67  & 0.57  & 0.52    & 0.8     \\
$r=3$   & Error    & 3.3   & \textbf{3.3}   & 3.34  & \textbf{3.3}$^*$   & 6.76 & 3.32  & \textbf{3.3}$^*$   & 3.4     & \textbf{3.3}$^*$     \\
$k=2$   & Sparsity & 2.2  & 2     & 1.85  & 2     & 1.6  & 1.99  & 1.99  & 1.83    & 2       \\ \hline
Urban   & Time     & 5.08  & 4.31  & 4.86  & 7.79  & 3.38  & 958   & 16.4  & 33.5    & 73.1    \\
$r=6$   & Error    & 7.67  & 8.13  & 8.62  & 7.83$^*$  & 8.97  & 8.07  & \textbf{7.76}$^*$  & 8.27    & \textbf{7.71}$^*$    \\ 
$k=2$   & Sparsity & 2.63  & 2     & 1.9   & 2     & 1.7   & 2     & 2     & 1.83    & 2       \\ \hline
Cuprite & Time     & 5.19  & 3.32  & 7.86  & 10.1  & 5.06  & 620   & 31.5  & 784     & 4829    \\
$r=12$  & Error    & 1.74  & 3.17  & 2.37  & 2.01  & 2.32  & 1.97  & \textbf{1.89}$^*$  & 1.93    & \textbf{1.83}$^*$    \\
$k=4$   & Sparsity & 6.61   & 4     & 3.92  & 4     & 3.53  & 4     & 4     & 3.81    & 4      
\end{tabular}}
\end{table}
We first note the natural sparsity of the data: without sparsity constraint, AS already produces very sparse solutions, and column-wise methods produce solutions with an average sparsity below the sparsity target $k$, meaning that some columns are naturally sparser than $k$.
We observe that the column-wise methods give relatively bad results in terms of reconstruction error, while the variants of Salmon are able to enforce sparsity while limiting the loss in error.
H+S is only slightly slower than Hcw, meaning that the selection (step 2) takes less time than the homotopy (step 1).
On the other hand, OG+S and ARBO+S are slower than their column-wise counterparts because they need to be run with a different sparsity target, respectively $k=r$ and $k=1$, to compute the whole Pareto front.
The computing times of H+S and OG+S seem proportional, and differ by a factor between $2$ and $3$, which is consistent with our discussion about computational cost in \cref{subsec:ouralgo}.
\lone{}-CD is very fast, and produces good solutions with some datasets (Jasper, Samson), but it is outperformed by Salmon in all cases and sometimes produces solutions with high error (CBCL, Cuprite).

Comparing OGg and OG+S, we observe that OGg is faster for tall matrices, while OG+S is faster for short and fat matrices.
Also, OG+S is always better in terms of reconstruction error, meaning that performing the heuristic column-wise and then recombining solutions is more efficient than applying the same heuristic matrix-wise. 
As regards Arborescent, we see a clear improvement of reconstruction error with Salmon, at the cost of a larger computation time.
The two-step approach of Salmon allows Arborescent to be applied for matrix-wise $q$-sparse MNNLS, which would be impossible otherwise.
Another advantage of the matrix-wise formulation is the capacity to tune the sparsity parameter more finely; this is illustrated on Jasper with $q/n=1.8$, for which we reach an average sparsity similar to that of column-wise metods and still get lower errors.
For the three variants of Salmon, the stars indicates that the selection (step 2) is done optimally as discussed in \cref{th:nearopti}; here it is the case for 19 out of 22 settings.

The abundance maps corresponding to the faces dataset Kuls are shown in \cref{fig:maps-jasper}; the other abundance maps from our experiments are available in \cref{app:suppxp}.
\begin{figure}[ht]
\captionsetup[subfloat]{farskip=1pt,captionskip=0.5pt}
\centering
\subfloat[AS (no sparsity constraint)]{
    \includegraphics[width=0.8\textwidth]{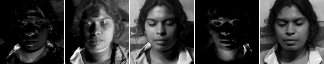}
}\\
\subfloat[\lone{}-CD]{
    \includegraphics[width=0.8\textwidth]{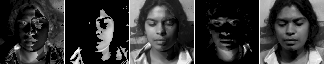}
}\\
\subfloat[Hcw]{
    \includegraphics[width=0.8\textwidth]{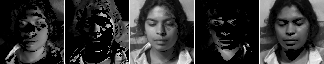}
}\\
\subfloat[H+S]{
    \includegraphics[width=0.8\textwidth]{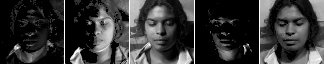}
}
  \caption{(1/2) Abundance maps (that is, reshaped rows of $X$) from the unmixing of the faces dataset Kuls by different algorithms.}
\end{figure}
\begin{figure}[ht]
\ContinuedFloat
\captionsetup[subfloat]{farskip=1pt,captionskip=0.5pt}
\centering
\subfloat[OGcw]{
    \includegraphics[width=0.8\textwidth]{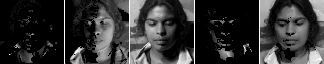}
}\\
\subfloat[OGg]{
    \includegraphics[width=0.8\textwidth]{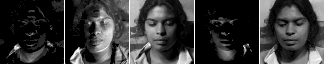}
}\\
\subfloat[OG+S]{
    \includegraphics[width=0.8\textwidth]{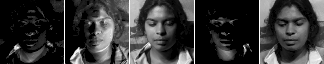}
}\\
\subfloat[ARBOcw]{
    \includegraphics[width=0.8\textwidth]{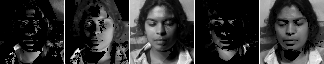}
}\\
\subfloat[ARBO+S]{
    \includegraphics[width=0.8\textwidth]{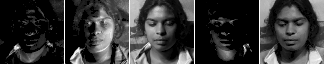}
}\\
  \caption{(2/2) Abundance maps (that is, reshaped rows of $X$) from the unmixing of the faces dataset Kuls by different algorithms.}
  \label{fig:maps-jasper}
\end{figure}
The features extracted in \cref{fig:maps-jasper} correspond to different directions of light.
Without sparsity constraint, the quality of the images is good, but the features are not well separated.
Column-wise methods and \lone{}-CD fail to retrieve the features and produce noisier images with pixelated regions.
Salmon, using any of the 3 possible methods for step 1, produces better-separated features, with a better spatial coherence.

\revise{
\subsection{Experiment 2: evolution of the computing time when $q$ varies}

In this experiment, we study the impact of the sparsity parameter $q$ on the running time of the matrix-wise sparse MNNLS algorithms.
For each setting, we run each algorithm 10 times and keep the minimum running time.
All algorithms are deterministic, so for a given setting the number of operations does not vary from one run to another and the differences in running time are due to the operating system, therefore taking the minimum time among several runs is a robust measure.
We also show the running time of the non-sparse method AS as a baseline.

\begin{figure}[p]
    \centering
    \includegraphics[width=0.75\textwidth]{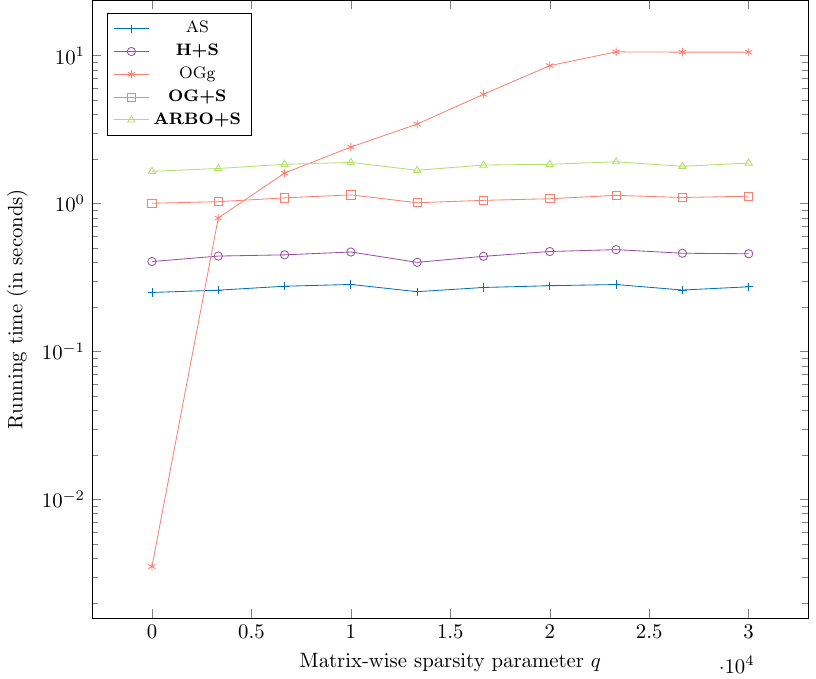}
    \caption{Evolution of the computing time of matrix-wise sparse NNLS algorithms, when $q$ varies, for the unmixing of the hyperspectral image Jasper.}
    \label{fig:xp-q-jasper}
\end{figure}
\begin{figure}[p]
    \centering
    \includegraphics[width=0.75\textwidth]{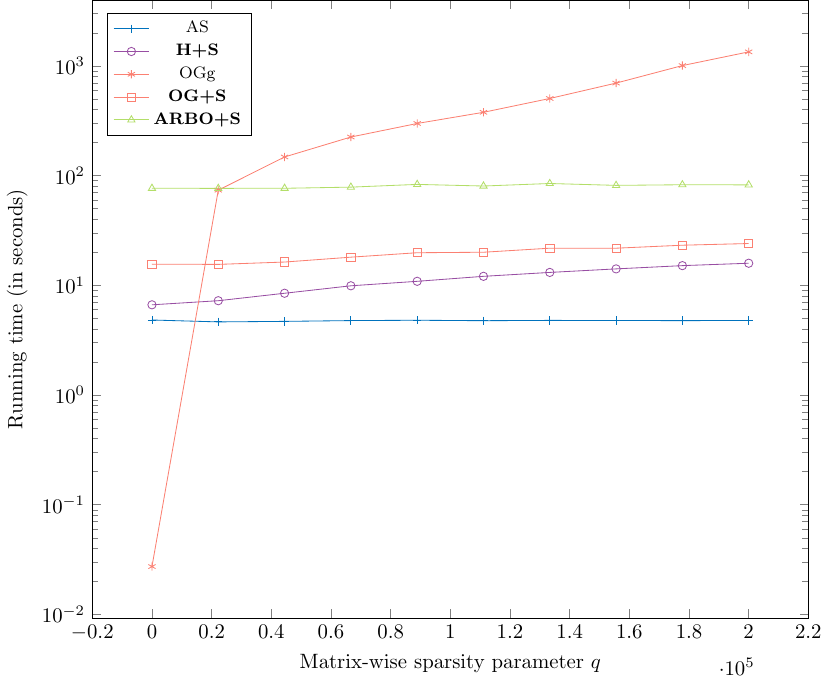}
    \caption{Evolution of the computing time of matrix-wise sparse NNLS algorithms, when $q$ varies, for the unmixing of the hyperspectral image Urban.}
    \label{fig:xp-q-urban}
\end{figure}

We consider the unmixing of the hyperspectral images Jasper (\cref{fig:xp-q-jasper}) and Urban (\cref{fig:xp-q-urban}) for different values of $q$.
On Jasper, the variants of Salmon have an almost constant running time, meaning that the cost of the selection step (step 2)is negligible compared to the cost of generating the Pareto fronts (step 1).
On the other hand, the running time of OGg grows exponentially (note the log scale of the vertical axis), although it is faster than Salmon when  $q$ is small. 
On Urban, the results are very similar.
The behaviour of ARBO+S and OGg is the same.
For H+S, and to a lesser extent OG+S, the computing time slightly grows as $q$ grows, meaning that the cost of the selection step is not negligible, but it is still dominated by the cost of step 1.

In this experiment, we find similar results on Jasper, a small image with $r=4$ where the column-wise subproblems of step 1 are very small, and Urban, a large image with $r=12$ where these subproblems are quite large.
This means that our selection algorithm is very fast and that the parameter $q$ does not increase significantly the computing time of Salmon.
}

\revise{
\subsection{Experiment 3: recovery of underlying solution on synthetic datasets}

In this experiment, we study the ability of the matrix-wise sparse MNNLS algorithms to recover the true solution in synthetic data sets where $X$ is generated with columns of different sparsities.
First, we generate $A \in \mathbb{R}^{100 \times 6}$ following the uniform distribution in $[0,1]$.
If we want $A$ to be ill-conditioned, we then compute the SVD of \mbox{$A=U \Sigma V^{\top}$}, replace the diagonal entries of $\Sigma$ by values between $10^{-4}$ and $1$ equally spaced in a log scale, and reconstruct \mbox{$A=U \Sigma V^{\top}$}.
Next, we generate $X \in \mathbb{R}^{6 \times 200}$ such that all columns are $k$-sparse with $k \in \{2,3,4\}$ chosen uniformly  at random, while the nonzero entries are generated uniformly at random in the interval [0,1].
We then compute $B=AX$. 
For the noise to be added to $B$, we first generate a matrix $N$ in which each entry is drawn from the normal distribution of mean $0$ and variance $1$, then rescale $N \gets \epsilon \frac{N}{\|N\|_2} \|B\|_2$ so that $\|N\|_2 = \epsilon \|B\|_2$, where $\epsilon$ is the noise level.

We generate this way 10 well-conditioned datasets and 10 ill-conditioned, and for different values of $\epsilon$ we generate and add noise to $B$. 
We then try to compute $X$ with different algorithms, given $A$ and the noisy $B$, and we measure the recovery rate, defined as the proportion of entries of the computed $X$ having the same value (in the sense zero or non-zero) than the corresponding entry of the generated $X$.
For each setting, we then average the recovery rate over the 10 datasets.
\Cref{fig:xp-synth-well,fig:xp-synth-ill} show the results for the well-conditioned and the ill-conditioned data, respectively.

\begin{figure}[p]
    \centering
    \includegraphics[width=0.75\textwidth]{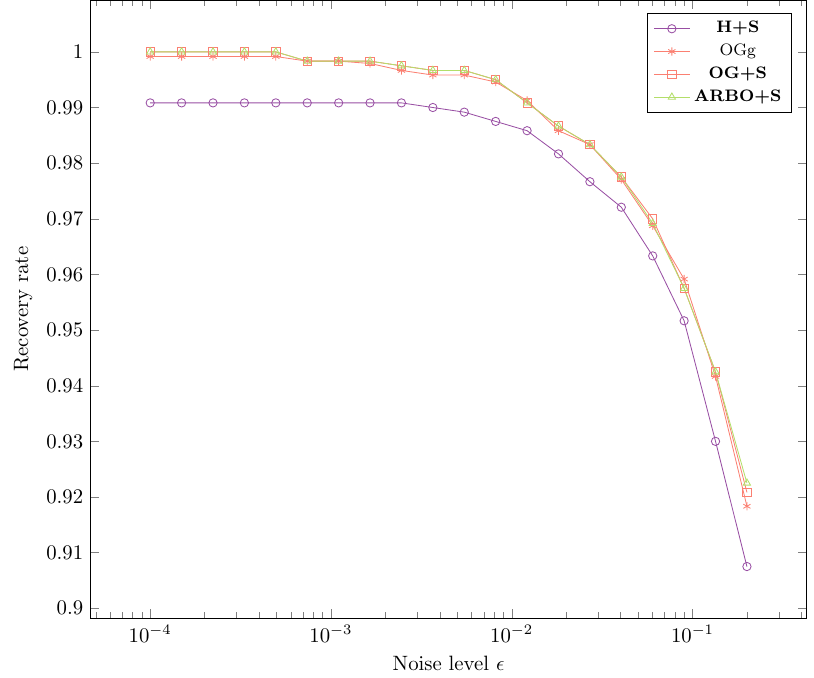}
    \caption{Evolution of the proportion of entries correctly recovered by matrix-wise sparse NNLS algorithms, on a synthetic well-conditioned dataset, when the noise level varies. The rate plotted is the average over 10 runs.}
    \label{fig:xp-synth-well}
\end{figure}
\begin{figure}[p]
    \centering
    \includegraphics[width=0.75\textwidth]{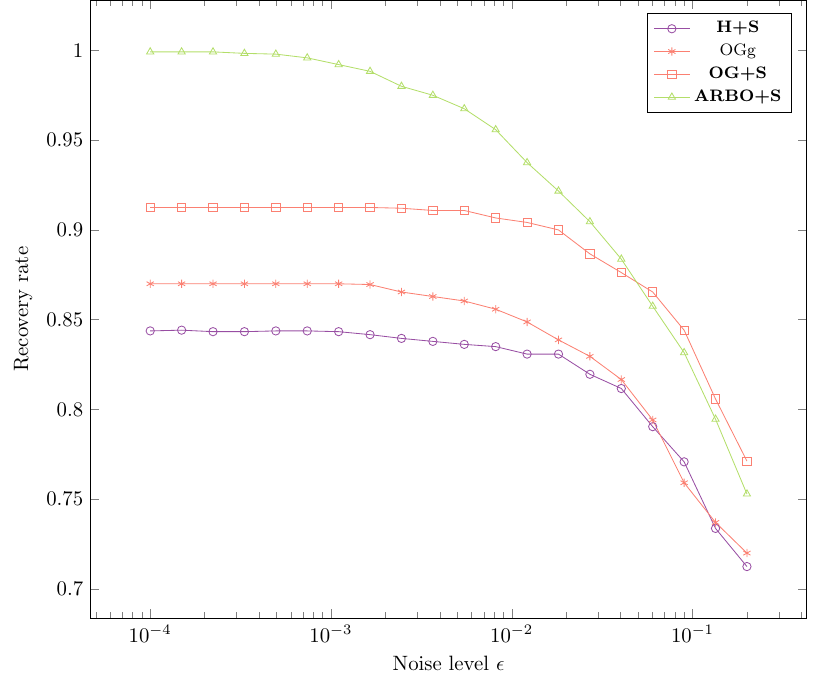}
    \caption{Evolution of the proportion of entries correctly recovered by matrix-wise sparse NNLS algorithms, on a synthetic ill-conditioned dataset, when the noise level varies. The rate plotted is the average over 10 runs.}
    \label{fig:xp-synth-ill}
\end{figure}

Surprisingly, for the well-conditioned dataset, all matrix-wise algorithms perform similarly, expect H+S that performs slightly worst.
When the noise level is below $10^{-2} = 1\%$, they almost perfectly recover the supports of the columns of $X$.
For higher noise levels, the recovery rate drops rapidly.  

For the ill-conditioned dataset, ARBO+S (the variant of Salmon using the exact algorithm Arborescent in step 1) has a recovery rate close to $100\%$ for noise smaller than $0.1\%$, while other variants do not perform as well.
The recovery rate of ARBO+S drops for noise above $0.1\%$ but it still performs better than the other variants of Salmon, until $1\%$ noise where OG+S becomes competitive.

As a conclusion, we see that using Arborescent in step 1 is especially effective with ill-conditioned data, which is often the case in real-world settings.
OG+S is also a good solution when the noise level is higher.
Using Arborescent is not very interesting when data is well-conditioned.
}

\paragraph{What algorithm should one use for step 1 of Salmon?}
When the dimension $r$ is small or when the computing time is not critical, the best option is to use Arborescent in step 1 as it is the only algorithm to compute the Pareto fronts exactly.
\revise{It is also the only algorithm to properly handle ill-conditioned data.}
In other cases, using NNOMP in step 1 generally produces better solutions than using the homotopy algorithm.
The homotopy algorithm is the fastest so it is appropriate when processing very large datasets with a limited time.
\revise{Note that there exist many other sparse NNLS algorithms that could be adapted to perform step 1, see for example \cite{mohimani2007fast} or \cite{blumensath2009iterative}.
We do not detail them for the sake of conciseness, and because our contribution lies not in the column-wise sparse methods but rather in how to use them in the two-step algorithm Salmon.}

\section{Conclusion}
\label{sec:conclu}

In this paper, we focused on the multiple nonnegative least squares problem with a matrix-wise \lzero{} constraint.
We introduced Salmon (\cref{alg:salmon}), that first computes for each column a Pareto front (step 1) and then applies a provably near-optimal selection strategy to build a solution matrix $X$ (step 2). 
We computed the Pareto fronts with three existing algorithms: one exact but slow branch-and-bound algorithm, and two fast heuristics.
We illustrated the advantages of Salmon for the unmixing of real-world facial and hyperspectral images, for which it outperformed state-of-the-art methods.
\revise{We also compared the different variants of Salmon on real-world and synthetic data sets to highlight their advantages and drawbacks.}

\begin{acknowledgements}
We thank Maxime De Wolf for his help in the implementation of the homotopy algorithm.
We thank T.T.\ Nguyen and her co-authors for making the codes of nonnegative greedy algorithms available online under a free software license. 
Finally, we thank the reviewers of this paper, whose comments helped significantly to improve the paper.
\end{acknowledgements}

\section*{Declarations}
{\small
\noindent \textbf{Funding}
NN and NG acknowledge the support by the European Research Council (ERC starting grant No 679515), and by the Fonds de la Recherche Scientifique - FNRS and the Fonds Wetenschappelijk Onderzoek - Vlanderen (FWO) under EOS project O005318F-RG47.
NG also acknowledges the Francqui foundation.
JEC acknowledges the support of the ANR grant ANR JCJC LoRAiA ANR-20-CE23-0010.

\vspace{1em}
\noindent \textbf{Conflicts of interest}
The authors have no conflict of interest or competing interest to declare.

\vspace{1em}
\noindent \textbf{Ethics approval, Consent to participate, Consent for publication}
Not applicable.

\vspace{1em}
\noindent \textbf{Availability of data, material and code}
The primary sources of the datasets used in our experiments are indicated in the main document.
We also include them along with our code and test scripts in the following online repository \url{https://gitlab.com/nnadisic/giant.jl}.

\vspace{1em}
\noindent \textbf{Authors' contributions}
NN designed the algorithm Salmon, implemented it, performed the experiments, and wrote the majority of the article.
JEC, AV, and NG all contributed significantly to the design of the algorithm and to the writing.
JEC provided expertise in greedy algorithms and helped with the implementation of the matrix-wise variant of NNOMP.
AV provided expertise on the homotopy algorithm and helped with its analysis and implementation.
NG supervised the work, significantly improved the proof of near-optimality, and revised the manuscript.

}

\bibliographystyle{spbasic}
\bibliography{bib.bib}

\appendix
\clearpage{}\section{The homotopy algorithm}
\label{app:homotopy}
In this section, we detailed the homotopy algorithm mentioned in \cref{subsec:homotopy}.

Given an $\ell_1$-NNLS problem, the homotopy algorithm computes sequentially all optimal solutions for the different values of $\lambda$. 
In a nutshell, it uses the KKT conditions (necessary conditions for optimality) to first find  the value $\lambda_{\max}$ for which the optimal solution of $\ell_1$-NNLS is $x=0$ for any $\lambda \geq \lambda_{\max}$, and then to compute the next smaller values of $\lambda$ for which the support (that is, the set of non-zero entries) of the optimal solution changes (one zero entry becomes non-zero, or the other way around). 
This is similar in spirit to active-set methods, akin to the simplex algorithm for linear programming. 
These values of $\lambda$ are called \emph{breakpoints}, between which the support of the optimal solution does not change.
We denote the first breakpoint $\lambda_{\max}$, and the following ones 
$\lambda_2$, $\lambda_3$, $\ldots$; 
see \cref{fig:path,fig:pareto} for an illustration.

\subsection{Optimality conditions}

The homotopy algorithm uses the first-order optimality conditions, that is, the KKT conditions, to determine the breakpoints and the supports of the corresponding solutions. 
Because $x$ is nonnegative, we have $\|x\|_1 = \sum_i x_i = e^T x$, where $e$ is the vector of all ones whose dimension will be clear from context. 
Therefore, the $\ell_1$-NNLS problem can be written as follows 
\begin{equation}
    \min\limits_{x \geq 0} f(x), \text{ where } f(x) = \frac{1}{2} \| Ax - b \|_2^2 + \lambda e^T x.
\end{equation} 
The KKT conditions are 
$x  \geq 0$,  
\begin{align}
    \nabla f(x)  = \underbrace{A^\top A}_{P} x - \underbrace{A^\top b}_{\ell} + \lambda e & \leq 0 , \label{eq:gradsup0} \\
     x_i (A^\top A x - A^\top b + \lambda e )_i & = 0 \text{ for all } i.  
     \label{eq:complcond}   
\end{align} 
\cref{eq:complcond} is the complementary condition; for every entry of $x$, either the entry itself or the corresponding gradient entry is equal to zero.
To simplify the notation, let us define $P = A^\top A$ and $\ell = A^\top b$. 

As $f$ is a convex function and the feasible set contains a Slater point (e.g., $x=e$), the KKT conditions are necessary and  sufficient. 
Therefore, any solution $x$ satisfying them is optimal. 
Suppose we know the optimal support, that is, the set $\mathcal{K}$ such that $x(\mathcal{K}) > 0$ and $x(\bar{\mathcal{K}}) = 0$, where $\bar{\mathcal{K}} = \{ 1,2,\ldots,n \} \setminus \mathcal{K} $.
Then 
\begin{align}
    x(\mathcal{K}) > 0 &\Rightarrow P(\mathcal{K},\mathcal{K})x(\mathcal{K}) - \ell(\mathcal{K}) + \lambda e = 0 \nonumber\\
    &\Rightarrow x(\mathcal{K}) = P(\mathcal{K},\mathcal{K})^{-1} (\ell(\mathcal{K}) - \lambda e) \geq 0, \tag{$\mathcal{C}1$} \label{eq:opticond1}
\end{align}
and
\begin{equation}
    P x - \ell + \lambda e \geq 0 \Rightarrow P(\bar{\mathcal{K}},\mathcal{K}) x(\mathcal{K}) - \ell(\bar{\mathcal{K}}) + \lambda e \geq 0 . \tag{$\mathcal{C}2$} \label{eq:opticond2}
\end{equation}
Replacing $x(\mathcal{K})$ in (\ref{eq:opticond2}) by (\ref{eq:opticond1}), we have
\begin{equation*}
    P(\bar{\mathcal{K}},\mathcal{K}) [ P(\mathcal{K},\mathcal{K})^{-1} (\ell(\mathcal{K}) - \lambda e) ] - \ell(\bar{\mathcal{K}}) + \lambda e \geq 0 .
\end{equation*}
Let us simplify the notation. Let
\begin{equation}
    a_\mathcal{K} = P(\mathcal{K},\mathcal{K})^{-1} \ell(\mathcal{K}), \text{ and }
    b_\mathcal{K} = P(\mathcal{K},\mathcal{K})^{-1} e. \label{eq:defak}
\end{equation}
We can rewrite (\ref{eq:opticond1}) as
\begin{equation*}\label{eq:shortopticond1}
    a_\mathcal{K} - \lambda b_\mathcal{K} \geq 0. \tag{$\mathcal{C}1b$}
\end{equation*}
Note that the dimension of $a_\mathcal{K}$ and $b_\mathcal{K}$ is the cardinality of $\mathcal{K}$. 
Let
\begin{equation*}
    c_\mathcal{K} = P(\bar{\mathcal{K}},\mathcal{K}) a_\mathcal{K} - \ell(\bar{\mathcal{K}}), \text{ and }
    d_\mathcal{K} = P(\bar{\mathcal{K}},\mathcal{K}) b_\mathcal{K} - e.
\end{equation*}
We can rewrite (\ref{eq:opticond2}) as
\begin{equation*}\label{eq:shortopticond2}
    c_\mathcal{K} - \lambda d_\mathcal{K} \geq 0. \tag{$\mathcal{C}2b$}
\end{equation*}
Note that the dimension of $c_\mathcal{K}$ and $d_\mathcal{K}$ is the cardinality of $\bar{\mathcal{K}}$. Moreover, Equations (\ref{eq:shortopticond1}) and (\ref{eq:shortopticond2}) are linear in $\lambda$: Given $\mathcal{K}$, we can easily compute $a_\mathcal{K}$, $b_\mathcal{K}$, $c_\mathcal{K}$, $d_\mathcal{K}$.

\subsection{Algorithm description} 

The goal of the homotopy algorithm is to compute breakpoints and their corresponding supports. 
It starts with an empty support, corresponding to the zero vector for any $\lambda \geq \lambda_{max}$, and iteratively adds or removes entries to the support while decreasing the value of $\lambda$.

The first step to build the regularization path is to find the first breakpoint $\lambda_{\max}$, that is, the minimum value of $\lambda$ such that the solution is the zero vector.
If the optimal solution is $x = 0$, that is, $\mathcal{K} = \varnothing$ and $\bar{\mathcal{K}} = \{ 1, 2, \ldots, n \}$, then from \cref{eq:gradsup0} we have $\lambda \geq \max\limits_i \ell_i$.
Therefore, the first breakpoint is 
\[ 
\lambda_{\max} = \max\limits_i \ell_i = \max\limits_i (A^\top b)_i =  \max\limits_i A(:,i)^T b. 
\] 
The index $i_{1} = \arg\max\limits_i \ell_i$ is the first to enter the support\footnote{If two of more columns maximize the value of $\ell_i$, we have to pick one to start the homotopy.  
If we always pick the one with smallest index, then the algorithm behaves normally, except that at the next iteration we will have $\lambda_{j+1} = \lambda_{j}$.
This is similar to Bland's rule for the simplex algorithm. }, thus for $\lambda_2 \leq \lambda < \lambda_{\max}$ we have $\mathcal{K} = \{ i_{1} \}$.

From a given support $\mathcal{K}_j$, the next breakpoint  $\lambda_{j+1} \leq \lambda_{j}$  is the largest value of $\lambda$ that violates one of the conditions (\ref{eq:opticond1}) or (\ref{eq:opticond2}). 
If (\ref{eq:opticond1}) is violated then a variable will leave the support, that is, a positive entry will become zero.
Denoting $k^*$ the index of this entry, we have $\mathcal{K}_{j+1} = \mathcal{K}_j \setminus \{ k^* \} $.
If (\ref{eq:opticond2}) is violated then a variable will enter the support, that is, a zero entry will become positive, $\mathcal{K}_{j+1} = \mathcal{K}_j \cup \{ k^* \} $.

Let us consider (\ref{eq:opticond1}).
We have $a_\mathcal{K}(k) - \lambda b_\mathcal{K}(k) \geq 0$ for all $k$ so
\begin{equation*}
    \lambda_{j+1} \geq \max_{\{k | b_\mathcal{K}(k) < 0\}} \frac{a_\mathcal{K}(k)}{b_\mathcal{K}(k)}.
\end{equation*}
Similarly, for (\ref{eq:opticond2}) we have  $c_\mathcal{K}(k) - \lambda d_\mathcal{K}(k) \geq 0$ for all $k$ so
\begin{equation*}
    \lambda_{j+1} \geq \max_{\{k | d_\mathcal{K}(k) < 0\}} \frac{c_\mathcal{K}(k)}{d_\mathcal{K}(k)}.
\end{equation*}
Therefore, 
\begin{equation*}\label{eq:lambdajplus1}
    \lambda_{j+1} = \max \Bigg( 
    \underbrace{\max_{\{k | b_\mathcal{K}(k) < 0\}} \frac{a_\mathcal{K}(k)}{b_\mathcal{K}(k)}}_{\text{Case } 1} , 
    \underbrace{\max_{\{k | d_\mathcal{K}(k) < 0\}} \frac{c_\mathcal{K}(k)}{d_\mathcal{K}(k)}}_{\text{Case } 2}
    \Bigg) . 
\end{equation*}
The algorithm is detailed formally in \cref{alg:homotopy}.
Note that, inside the algorithm loop, once a support $\mathcal{K}$ is identified, getting the corresponding optimal solution is straightforward. 
From \cref{eq:defak}, if $a_\mathcal{K}$ is nonnegative, then the unbiased optimal solution of the NNLS problem is the vector $x^*$ such that $x^*(\bar{\mathcal{K}}) = 0$ and $x^*(\mathcal{K}) = a_\mathcal{K}$.
If $a_\mathcal{K}$ has negative entries, then we can compute the unbiased solution with a standard NNLS solver, with $x^*(\bar{\mathcal{K}}) = 0$ and $x^*(\mathcal{K}) = \arg\min\limits_{x \geq 0} \| P(\mathcal{K},\mathcal{K}) x - l(\mathcal{K}) \|_2^2$.

\begin{algorithm}[htp]

  \KwIn{$A \in \mathbb{R}^{m \times r}$,  $b \in \mathbb{R}^m$}
  \KwOut{Breakpoints $\lambda_j$, corresponding supports $\mathcal{K}_j$ and solutions $x^*_j$, for all $j$}

  \DontPrintSemicolon

    $i_{1} \leftarrow \arg\max\limits_i \ell_i$\\
    $K \leftarrow \{ i_{1} \}$ ; $\lambda_1 = \lambda_{\max} = \ell_{i_{1}}$, $j \leftarrow 1$ \\ \While{ $\lambda_j > 0$ }
    {         
        $j \leftarrow j + 1$\\
        Compute $a_K$, $b_K$, $c_K$, $d_K$ \\
        $\lambda_{C1} \leftarrow \max_{\{k | b_\mathcal{K}(k) < 0\}} \frac{a_\mathcal{K}(k)}{b_\mathcal{K}(k)}$, and $k^*_{C1} \gets$ corresponding argmax\\
        $\lambda_{C2} \leftarrow \max_{\{k | d_\mathcal{K}(k) < 0\}} \frac{c_\mathcal{K}(k)}{d_K(k)}$, and $k^*_{C2} \gets$ corresponding argmax\\
\If{$\lambda_{C1} \geq \lambda_{C2}$} 
{ 
            $\lambda_{j} \leftarrow \lambda_{C1}$ \\
            $\mathcal{K} \leftarrow \mathcal{K} \setminus \{k^*_{C1}\}$
        }
\ElseIf{$\lambda_{C1} < \lambda_{C2}$}{
            $\lambda_{j} \leftarrow \lambda_{C2}$ \\
            $\mathcal{K} \leftarrow \mathcal{K} \cup      \{k^*_{C2}\}$
        }
        
        $x^*_j(\bar{\mathcal{K}}) \leftarrow 0$ \\
        \If{$a_\mathcal{K} > 0$}{ 
            $x^*_j(\mathcal{K}) \leftarrow a_K$
        }
        \Else{
$x^*_\mathcal{K} = \arg\min\limits_{x \geq 0} \| A(:,\mathcal{K}) x - b(\mathcal{K}) \|_2^2$
        }
    }
  
  \caption{Homotopy algorithm for sparse NNLS}
  \label{alg:homotopy}
\end{algorithm}

\subsection{Computational cost}
\label{sec:homotopy-analysis}

As explained by \cite{kim2013regularization}, the time complexity of one iteration of the homotopy algorithm is the same as one iteration of the standard active-set algorithm \citep{lawson1995solving}, that is, $\mathcal{O}(r^3)$ operations.
It is dominated by the computation of $a_\mathcal{K}$, that entails solving a linear system in at most $r$ variables. 
The number of iterations equals the number of breakpoints, which is in practice similar to those of the active-set. In the worst case, active-set methods might require an exponential number of iterations, up to $\mathcal{O}(2^r)$, as the simplex algorithm for linear programming.
However, in practice, we have observed that it typically requires much less iterations, of the order of $\mathcal{O}(r)$. 
In particular, when $P^{-1}$ is diagonally dominant, we have $b_\mathcal{K} > 0$, so (\ref{eq:opticond1}) is never violated. 
As a result, when $\lambda$ decreases, no positive entry of $x$ becomes zero. 
We only add entries to the support, so the homotopy algorithm will be done in at most $r$ iterations.
In practice, even when this condition is not met, we have observed that adding entries to the support happens far more often than removing entries. 

As the homotopy algorithm solves a series of $\ell_1$-penalized NNLS problems, there exist conditions under which it is guaranteed to recover the correct supports, that is, the supports of the solutions of the corresponding $\ell_0$-constrained NNLS problems; see \cite{itoh2017perfecta} and the references therein.
\clearpage{} 
\clearpage{}
\section{Additional experimental results}\label{app:suppxp}

In this document, we provide the abundance maps resulting from our experiments that could not be included in the paper:
\begin{itemize}
    \item CBCL in Figure~\ref{fig:supp-xpcbcl}, 
    \item Frey in Figure~\ref{fig:supp-xpfrey},  
\item Jasper in Figure~\ref{fig:supp-xpjasper}, 
    \item Samson in Figure~\ref{fig:supp-xpsamson},  
    \item Urban in Figure~\ref{fig:supp-xpurban}, and 
    \item Cuprite in Figure~\ref{fig:supp-xpcuprite}.  
\end{itemize}

\begin{figure}[ht]
\centering
\subfloat[AS]{
    \includegraphics[width=0.48\textwidth]{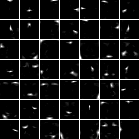}
}
\subfloat[\lone{}-CD]{
    \includegraphics[width=0.48\textwidth]{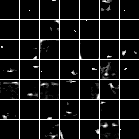}
}\\
\subfloat[Hcw]{
    \includegraphics[width=0.48\textwidth]{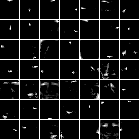}
}
\subfloat[H+S]{
    \includegraphics[width=0.48\textwidth]{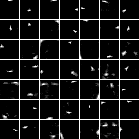}
}
  \caption{(1/2) Abundance maps (that is, reshaped rows of $X$) from the unmixing of the faces dataset CBCL by different algorithms.}
\end{figure}
\begin{figure}[ht]
\ContinuedFloat
\centering
\subfloat[OGcw]{
    \includegraphics[width=0.48\textwidth]{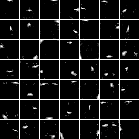}
}
\subfloat[OGg]{
    \includegraphics[width=0.48\textwidth]{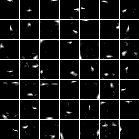}
}\\
\subfloat[OG+S]{
    \includegraphics[width=0.48\textwidth]{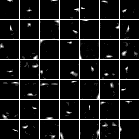}
}
  \caption{(2/2) Abundance maps (that is, reshaped rows of $X$) from the unmixing of the faces dataset CBCL by different algorithms.}
  \label{fig:supp-xpcbcl}
\end{figure}

\begin{figure}[ht]
\centering
\subfloat[AS]{
    \includegraphics[angle=270,width=0.47\textwidth]{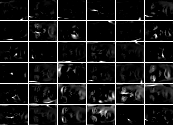}
}
\subfloat[\lone{}-CD]{
    \includegraphics[angle=270,width=0.47\textwidth]{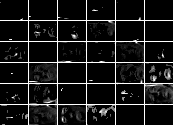}
}\\
\subfloat[Hcw]{
    \includegraphics[angle=270,width=0.47\textwidth]{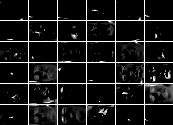}
}
\subfloat[H+S]{
    \includegraphics[angle=270,width=0.47\textwidth]{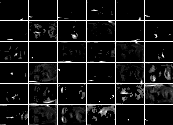}
}
  \caption{(1/2) Abundance maps (that is, reshaped rows of $X$) from the unmixing of the faces dataset Frey by different algorithms.}
\end{figure}
\begin{figure}[ht]
\ContinuedFloat
\centering
\subfloat[OGcw]{
    \includegraphics[angle=270,width=0.47\textwidth]{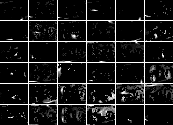}
}
\subfloat[OGg]{
    \includegraphics[angle=270,width=0.47\textwidth]{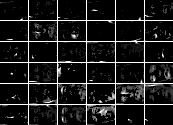}
}\\
\subfloat[OG+S]{
    \includegraphics[angle=270,width=0.47\textwidth]{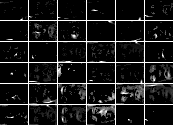}
}
  \caption{(2/2) Abundance maps (that is, reshaped rows of $X$) from the unmixing of the faces dataset Frey by different algorithms.}
  \label{fig:supp-xpfrey}
\end{figure}

\begin{figure}[ht]
\centering
\subfloat[AS]{
    \includegraphics[width=0.95\textwidth]{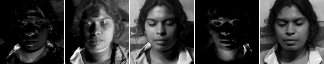}
}\\
\subfloat[\lone{}-CD]{
    \includegraphics[width=0.95\textwidth]{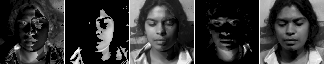}
}\\
\subfloat[Hcw]{
    \includegraphics[width=0.95\textwidth]{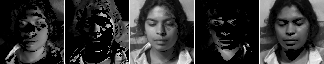}
}\\
\subfloat[H+S]{
    \includegraphics[width=0.95\textwidth]{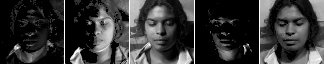}
}
  \caption{(1/2) Abundance maps (that is, reshaped rows of $X$) from the unmixing of the faces dataset Kuls by different algorithms.}
\end{figure}
\begin{figure}[ht]
\ContinuedFloat
\centering
\subfloat[OGcw]{
    \includegraphics[width=0.95\textwidth]{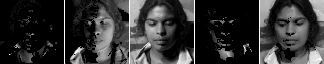}
}\\
\subfloat[OGg]{
    \includegraphics[width=0.95\textwidth]{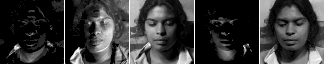}
}\\
\subfloat[OG+S]{
    \includegraphics[width=0.95\textwidth]{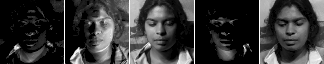}
}\\
\subfloat[ARBOcw]{
    \includegraphics[width=0.95\textwidth]{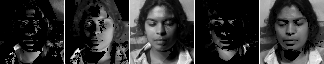}
}\\
\subfloat[ARBO+S]{
    \includegraphics[width=0.95\textwidth]{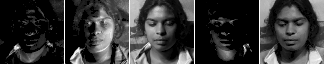}
}\\
  \caption{(2/2) Abundance maps (that is, reshaped rows of $X$) from the unmixing of the faces dataset Kuls by different algorithms.}
  \label{fig:supp-xpkuls}
\end{figure}

\begin{figure}[ht]
\centering
\subfloat[AS]{
    \includegraphics[width=0.90\textwidth]{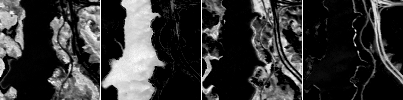}
}\\
\subfloat[\lone{}-CD]{
    \includegraphics[width=0.90\textwidth]{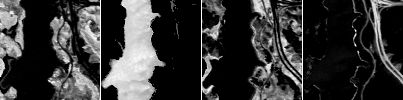}
}\\
\subfloat[Hcw]{
    \includegraphics[width=0.90\textwidth]{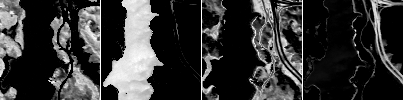}
}\\
\subfloat[H+S]{
    \includegraphics[width=0.90\textwidth]{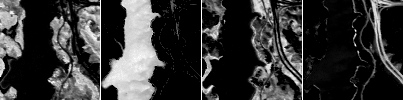}
}
  \caption{(1/2) Abundance maps (that is, reshaped rows of $X$) from the unmixing of the hyperspectral image Jasper by different algorithms.}
\end{figure}
\begin{figure}[ht]
\ContinuedFloat
\centering
\subfloat[OGcw]{
    \includegraphics[width=0.90\textwidth]{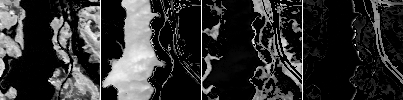}
}\\
\subfloat[OGg]{
    \includegraphics[width=0.90\textwidth]{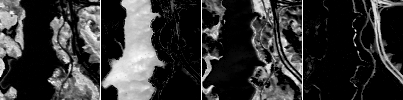}
}\\
\subfloat[OG+S]{
    \includegraphics[width=0.90\textwidth]{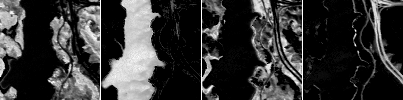}
}\\
\subfloat[ARBOcw]{
    \includegraphics[width=0.90\textwidth]{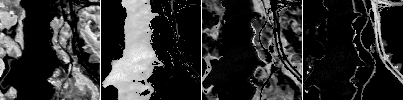}
}\\
\subfloat[ARBO+S]{
    \includegraphics[width=0.90\textwidth]{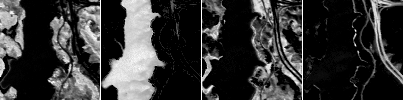}
}\\
  \caption{(2/2) Abundance maps (that is, reshaped rows of $X$) from the unmixing of the hyperspectral image Jasper by different algorithms.}
  \label{fig:supp-xpjasper}
\end{figure}

\begin{figure}[ht]
\centering
\subfloat[\lone{}-CD]{
    \includegraphics[width=0.90\textwidth]{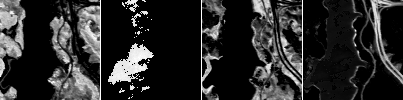}
}\\
\subfloat[H+S]{
    \includegraphics[width=0.90\textwidth]{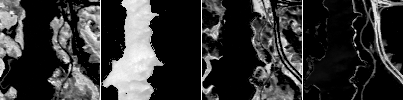}
}\\
\subfloat[OGg]{
    \includegraphics[width=0.90\textwidth]{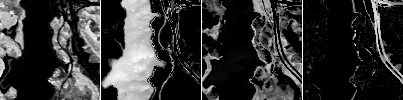}
}\\
\subfloat[OG+S]{
    \includegraphics[width=0.90\textwidth]{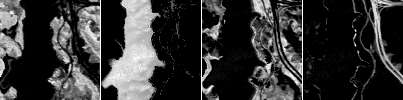}
}\\
\subfloat[ARBO+S]{
    \includegraphics[width=0.90\textwidth]{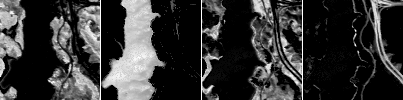}
}\\
  \caption{Abundance maps (that is, reshaped rows of $X$) from the unmixing of the hyperspectral image Jasper by different algorithms, with $k=q/n=1.8$.}
  \label{fig:supp-xpjasperk18}
\end{figure}

\begin{figure}[ht]
\centering
\subfloat[AS]{
    \includegraphics[width=0.70\textwidth]{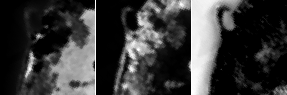}
}\\
\subfloat[\lone{}-CD]{
    \includegraphics[width=0.70\textwidth]{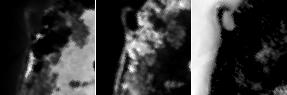}
}\\
\subfloat[Hcw]{
    \includegraphics[width=0.70\textwidth]{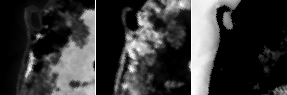}
}\\
\subfloat[H+S]{
    \includegraphics[width=0.70\textwidth]{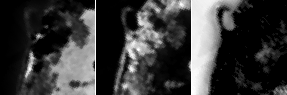}
}
  \caption{(1/2) Abundance maps (that is, reshaped rows of $X$) from the unmixing of the hyperspectral image Samson by different algorithms.}
\end{figure}
\begin{figure}[ht]
\ContinuedFloat
\centering
\subfloat[OGcw]{
    \includegraphics[width=0.70\textwidth]{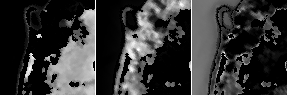}
}\\
\subfloat[OGg]{
    \includegraphics[width=0.70\textwidth]{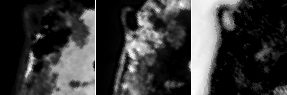}
}\\
\subfloat[OG+S]{
    \includegraphics[width=0.70\textwidth]{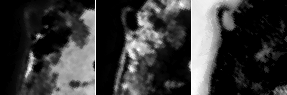}
}\\
\subfloat[ARBOcw]{
    \includegraphics[width=0.70\textwidth]{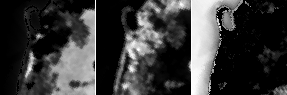}
}\\
\subfloat[ARBO+S]{
    \includegraphics[width=0.70\textwidth]{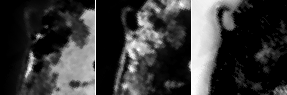}
}\\
  \caption{(2/2) Abundance maps (that is, reshaped rows of $X$) from the unmixing of the hyperspectral image Samson by different algorithms.}
  \label{fig:supp-xpsamson}
\end{figure}

\begin{figure}[ht]
\centering
\subfloat[AS]{
    \includegraphics[width=0.98\textwidth]{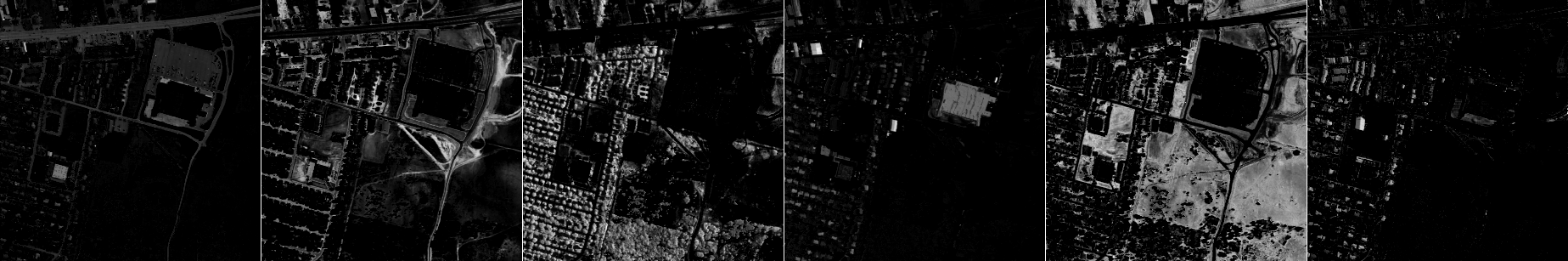}
}\\
\subfloat[\lone{}-CD]{
    \includegraphics[width=0.98\textwidth]{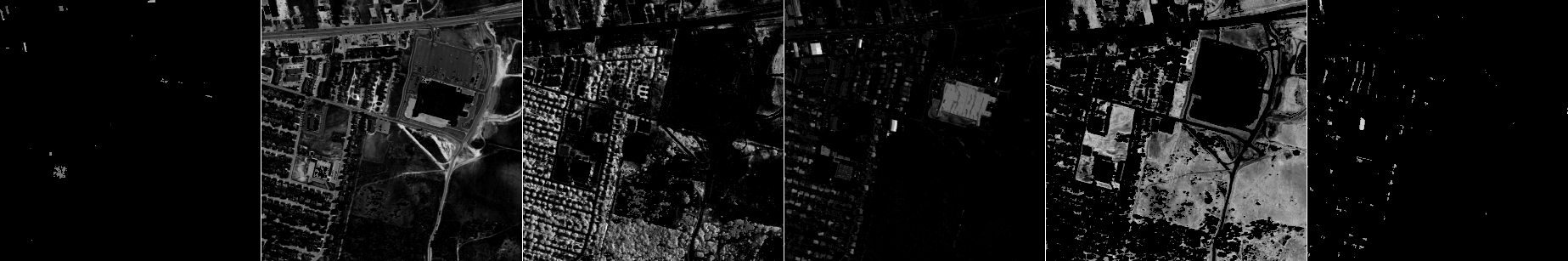}
}\\
\subfloat[Hcw]{
    \includegraphics[width=0.98\textwidth]{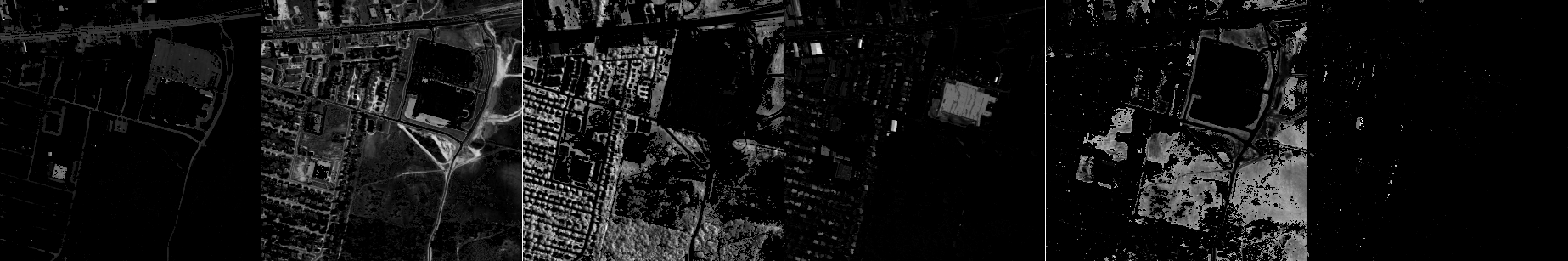}
}\\
\subfloat[H+S]{
    \includegraphics[width=0.98\textwidth]{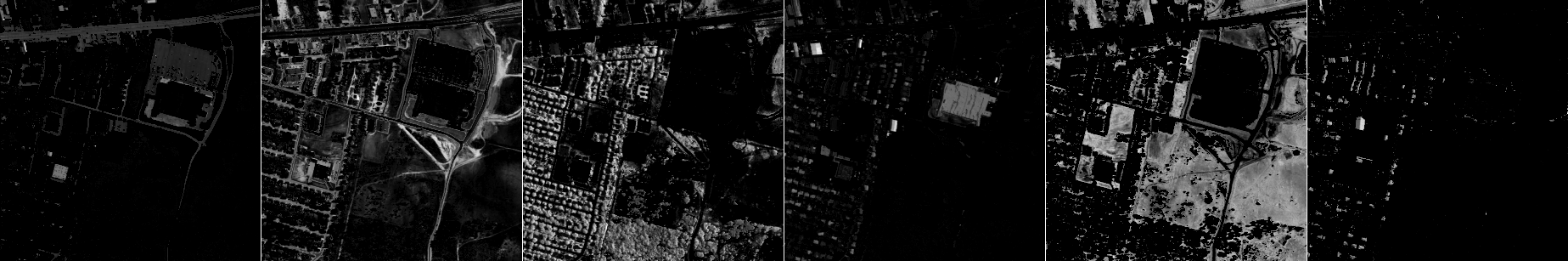}
}
  \caption{(1/2) Abundance maps (that is, reshaped rows of $X$) from the unmixing of the hyperspectral image Urban by different algorithms.}
\end{figure}
\begin{figure}[ht]
\ContinuedFloat
\centering
\subfloat[OGcw]{
    \includegraphics[width=0.98\textwidth]{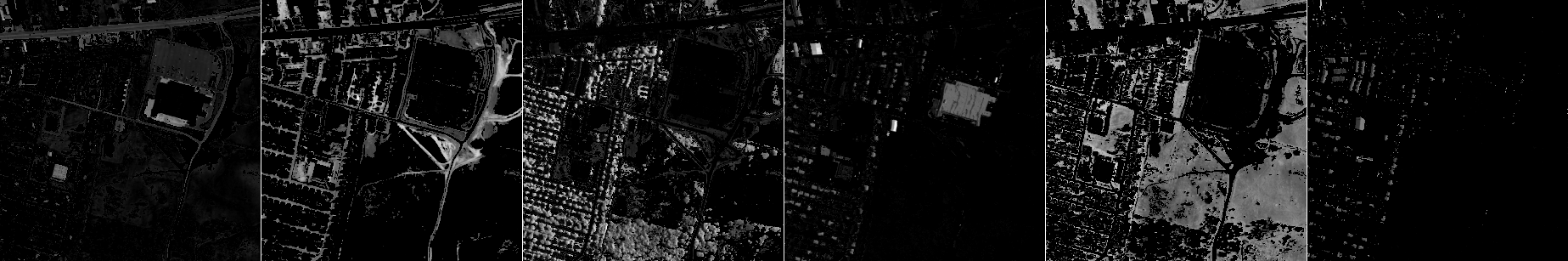}
}\\
\subfloat[OGg]{
    \includegraphics[width=0.98\textwidth]{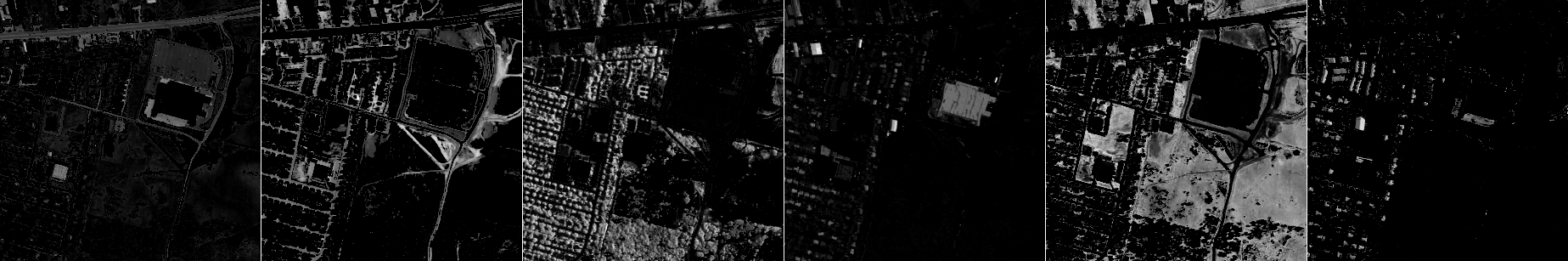}
}\\
\subfloat[OG+S]{
    \includegraphics[width=0.98\textwidth]{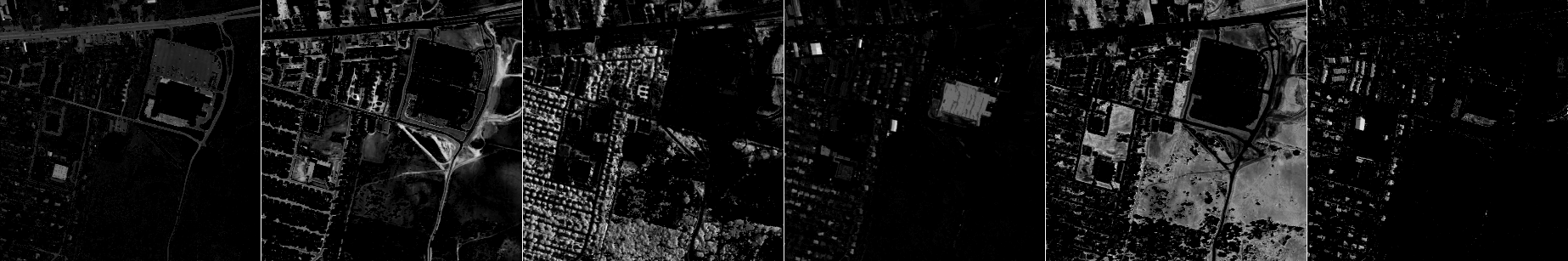}
}\\
\subfloat[ARBOcw]{
    \includegraphics[width=0.98\textwidth]{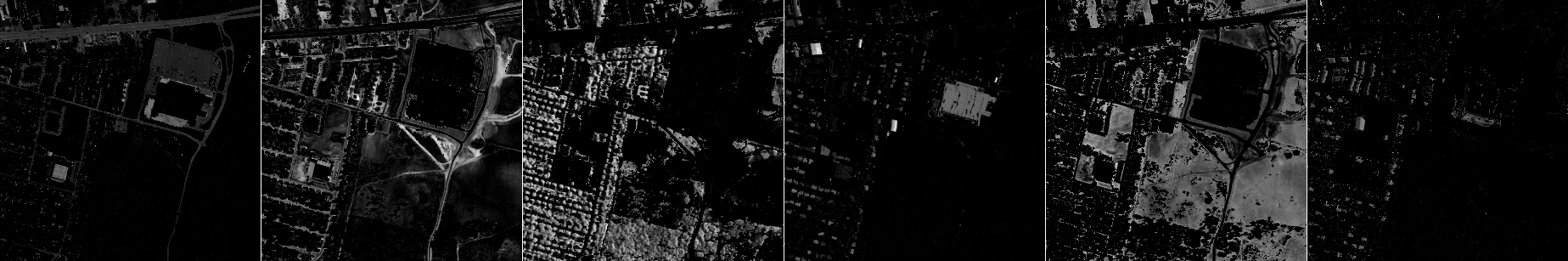}
}\\
\subfloat[ARBO+S]{
    \includegraphics[width=0.98\textwidth]{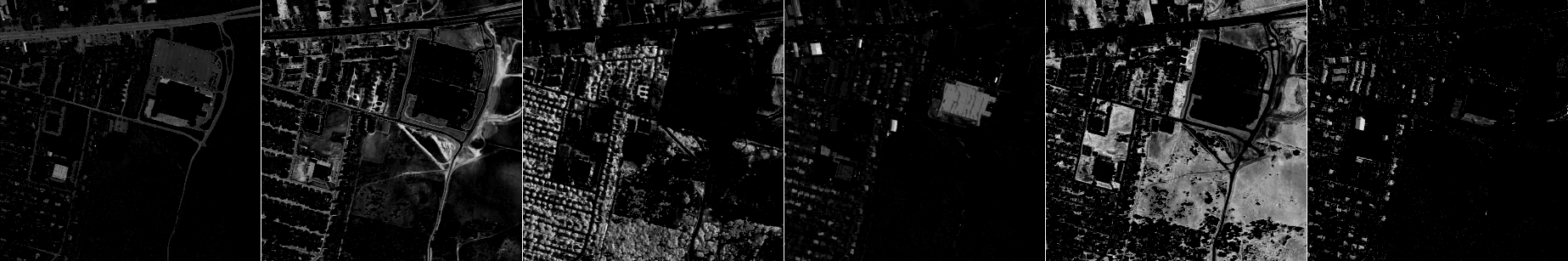}
}\\
  \caption{(2/2) Abundance maps (that is, reshaped rows of $X$) from the unmixing of the hyperspectral image Urban by different algorithms.}
  \label{fig:supp-xpurban}
\end{figure}

\begin{figure}[ht]
\centering
\subfloat[AS]{
    \includegraphics[width=0.45\textwidth]{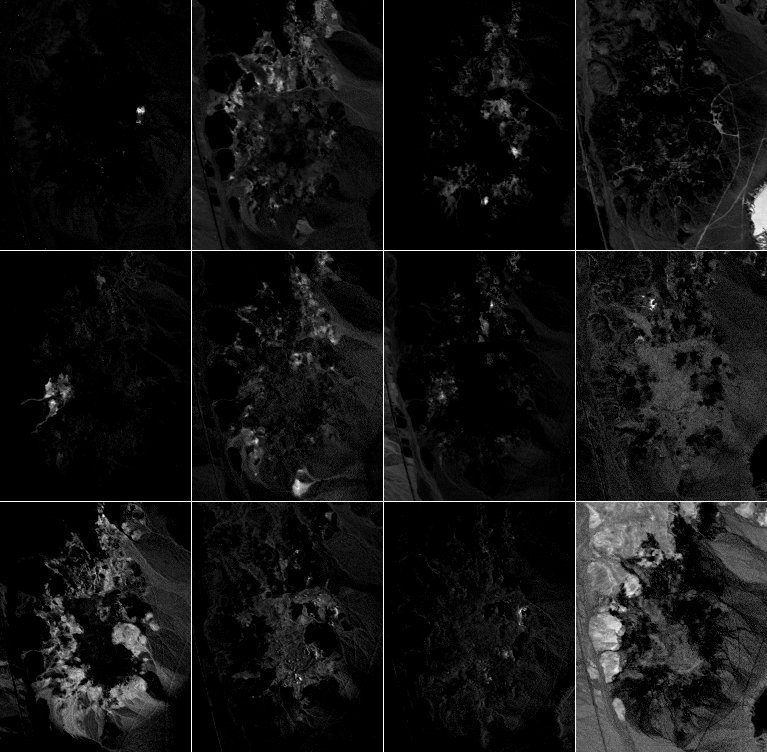}
}
\subfloat[\lone{}-CD]{
    \includegraphics[width=0.45\textwidth]{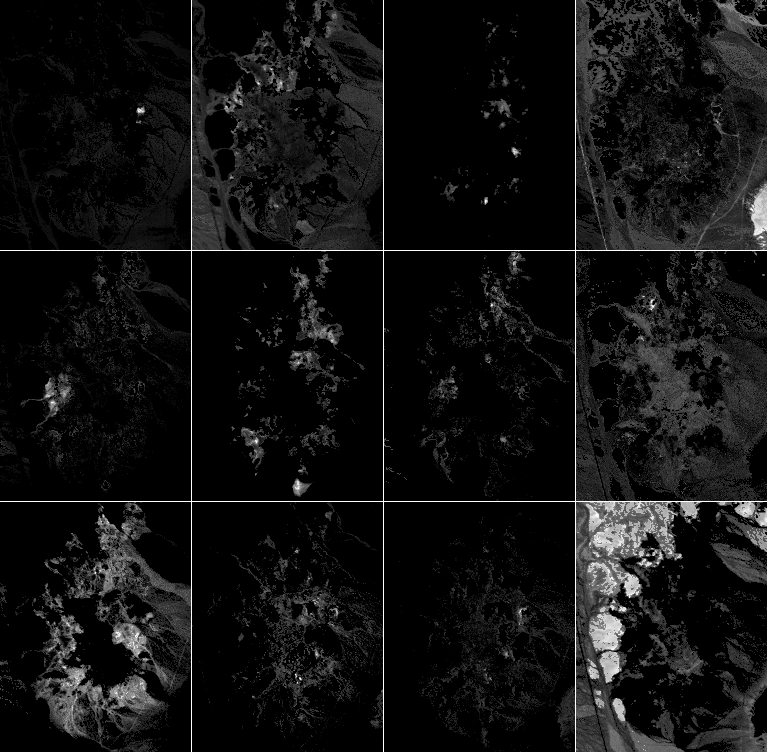}
}\\
\subfloat[Hcw]{
    \includegraphics[width=0.45\textwidth]{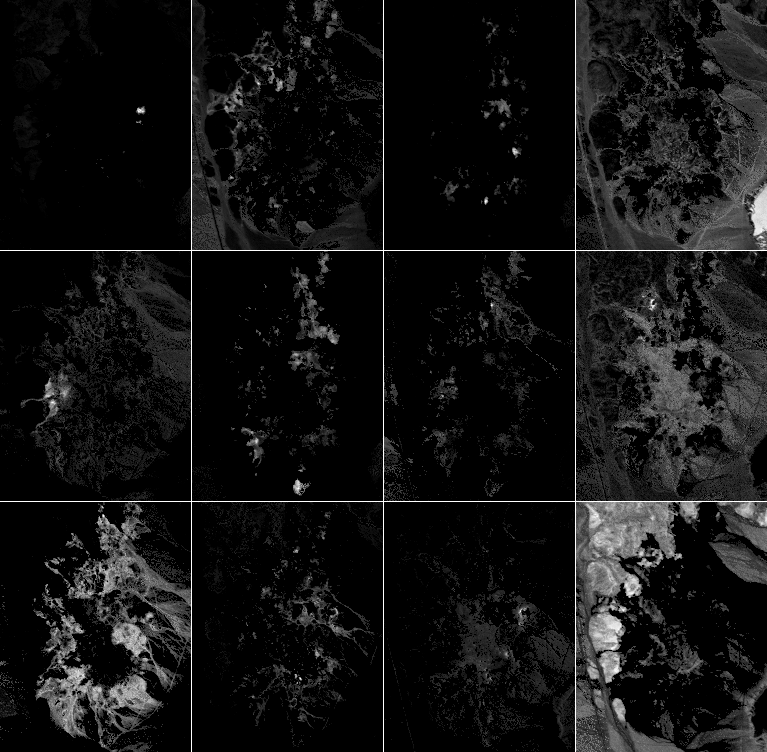}
}
\subfloat[H+S]{
    \includegraphics[width=0.45\textwidth]{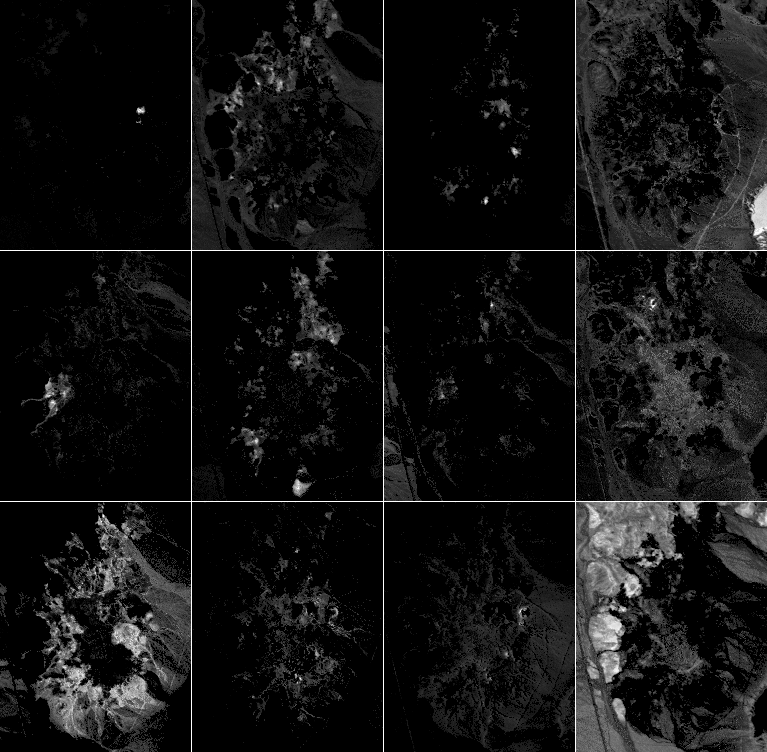}
}
  \caption{(1/2) Abundance maps (that is, reshaped rows of $X$) from the unmixing of the hyperspectral image Cuprite by different algorithms.}
\end{figure}
\begin{figure}[ht]
\ContinuedFloat
\centering
\subfloat[OGcw]{
    \includegraphics[width=0.45\textwidth]{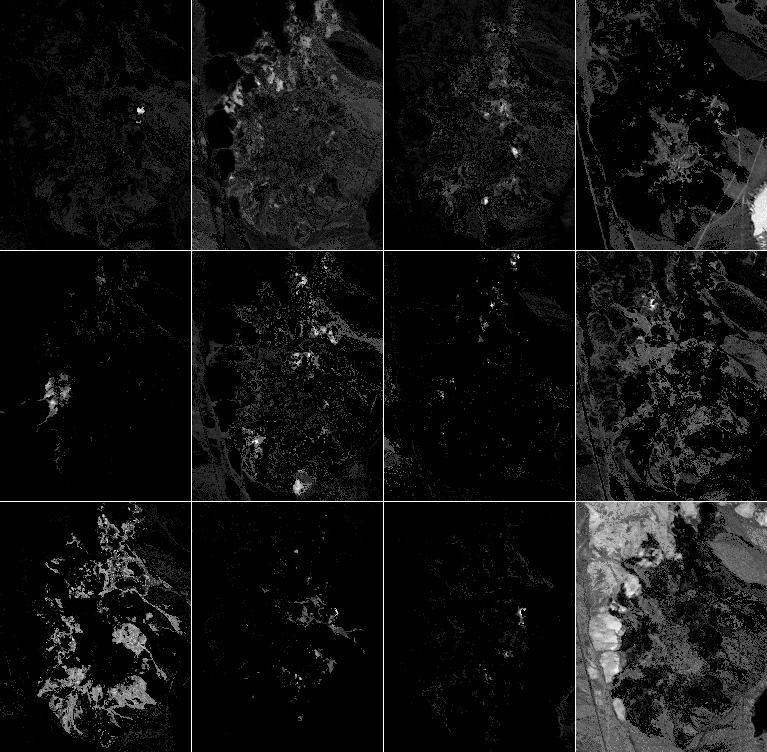}
}
\subfloat[OGg]{
    \includegraphics[width=0.45\textwidth]{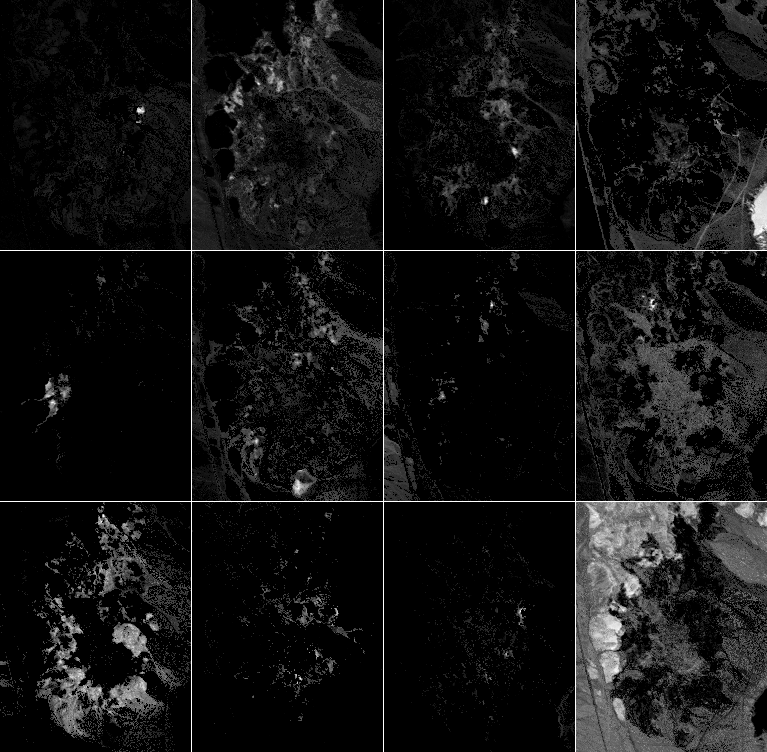}
}\\
\subfloat[OG+S]{
    \includegraphics[width=0.45\textwidth]{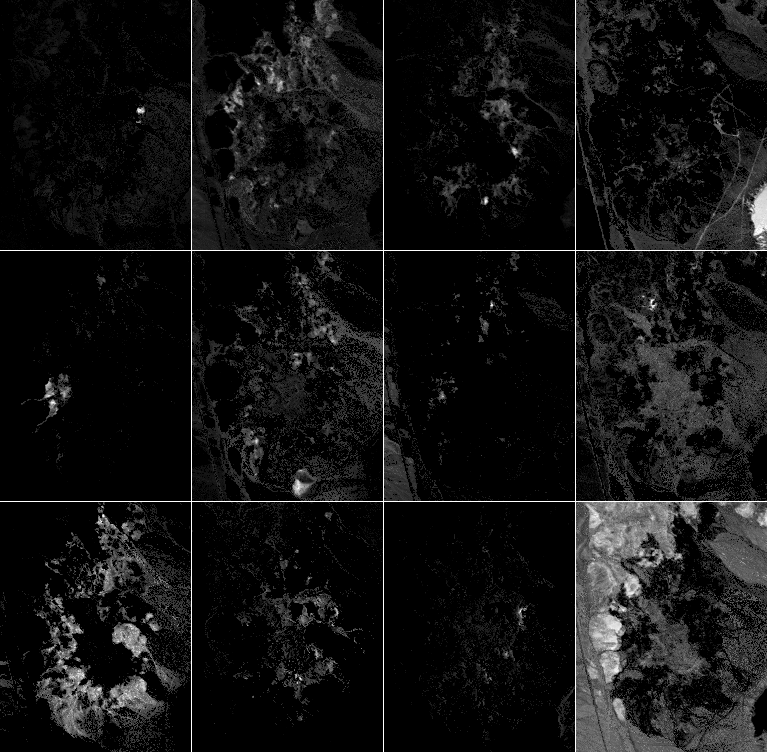}
}\\
\subfloat[ARBOcw]{
    \includegraphics[width=0.45\textwidth]{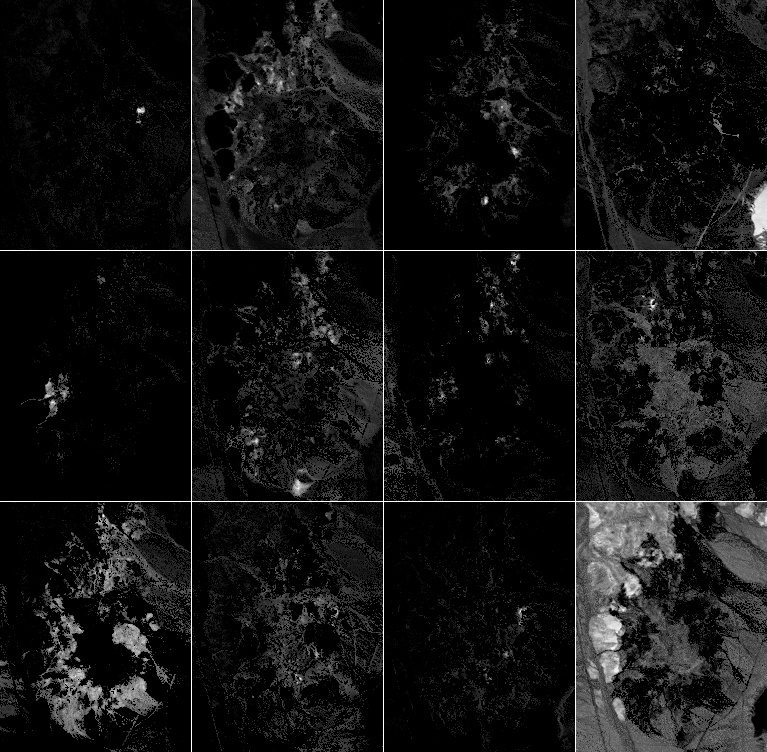}
}
\subfloat[ARBO+S]{
    \includegraphics[width=0.45\textwidth]{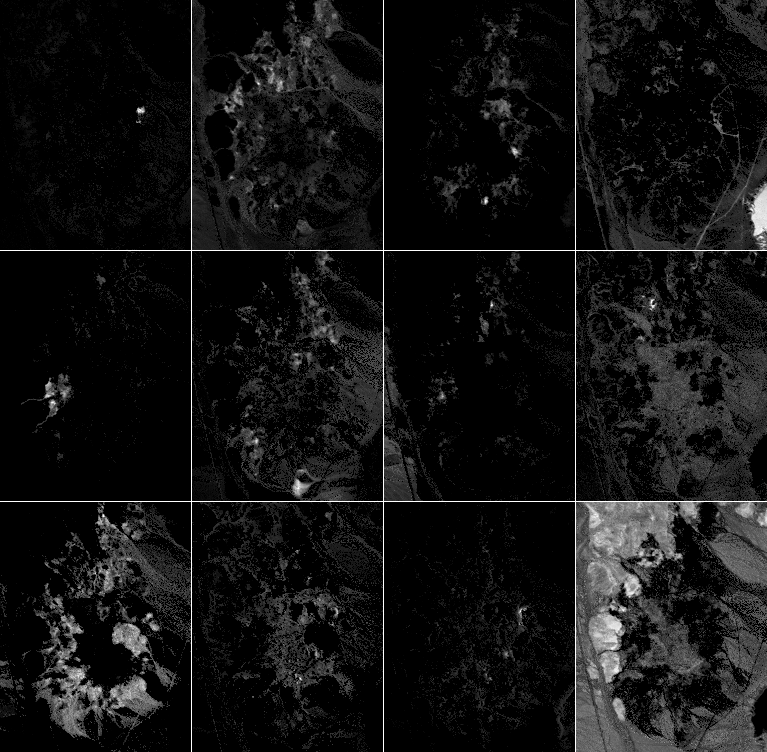}
}\\
  \caption{(2/2) Abundance maps (that is, reshaped rows of $X$) from the unmixing of the hyperspectral image Cuprite by different algorithms.}
  \label{fig:supp-xpcuprite}
\end{figure}
\clearpage{}

\end{document}